\documentclass[10pt,twocolumn,letterpaper]{article}

\usepackage[pagenumbers]{cvpr} %

\newcommand\blfootnote[1]{%
  \begingroup
  \renewcommand\thefootnote{}\footnote{#1}%
  \addtocounter{footnote}{-1}%
  \endgroup
}

\newcommand{\name}{LoST\xspace}
\newcommand{\webpage}{\href{http://niladridutt.com/lost}{webpage}\xspace}

\usepackage[normalem]{ulem}

\usepackage{microtype}

\definecolor{cvprblue}{rgb}{0.21,0.49,0.74}
\usepackage[pagebackref,breaklinks,colorlinks,allcolors=cvprblue]{hyperref}
\usepackage{booktabs}   
\usepackage{multirow}   
\usepackage{graphicx}  
\usepackage{mathtools}
\usepackage{amssymb}%
\usepackage{pifont}%
\def\confName{CVPR}
\def\confYear{2026}

\usepackage{tcolorbox}
\usepackage{xcolor}
\newtcolorbox{promptbox}[1][]{
  colback=gray!5,       %
  colframe=gray!40,     %
  arc=5pt,              %
  boxrule=0.8pt,        %
  width=\textwidth,     %
  title=\textbf{Text Prompt Template}, %
  fonttitle=\bfseries\small,
  coltitle=black,       %
  #1
}

\title{
LoST: Level of Semantics Tokenization for 3D Shapes}

\author{%
\fontsize{11pt}{13pt}\selectfont
Niladri Shekhar Dutt\textsuperscript{1,2*}\quad
Zifan Shi\textsuperscript{2} \quad
Paul Guerrero\textsuperscript{2} \quad
Chun-Hao Paul Huang\textsuperscript{2} \quad \\[0.5ex]
\fontsize{11pt}{13pt}\selectfont
Duygu Ceylan\textsuperscript{2} \quad 
Niloy J. Mitra\textsuperscript{1,2} \quad
Xuelin Chen\textsuperscript{2} \\[1.5ex]
\fontsize{11pt}{13pt}\selectfont
\textsuperscript{1}University College London \quad
\textsuperscript{2}Adobe Research 
\\ [1.5ex]
\fontsize{11pt}{13pt}\selectfont
\href{https://lost3d.github.io/}{https://lost3d.github.io}
}

\begin{document}

 \twocolumn[{%
 \renewcommand\twocolumn[1][]{#1}%
 \maketitle
 \vspace{-0.3in}
 \thispagestyle{empty}
 \begin{center}
 \centering 
 \captionsetup{type=figure}
 \includegraphics[width=1.0\linewidth]{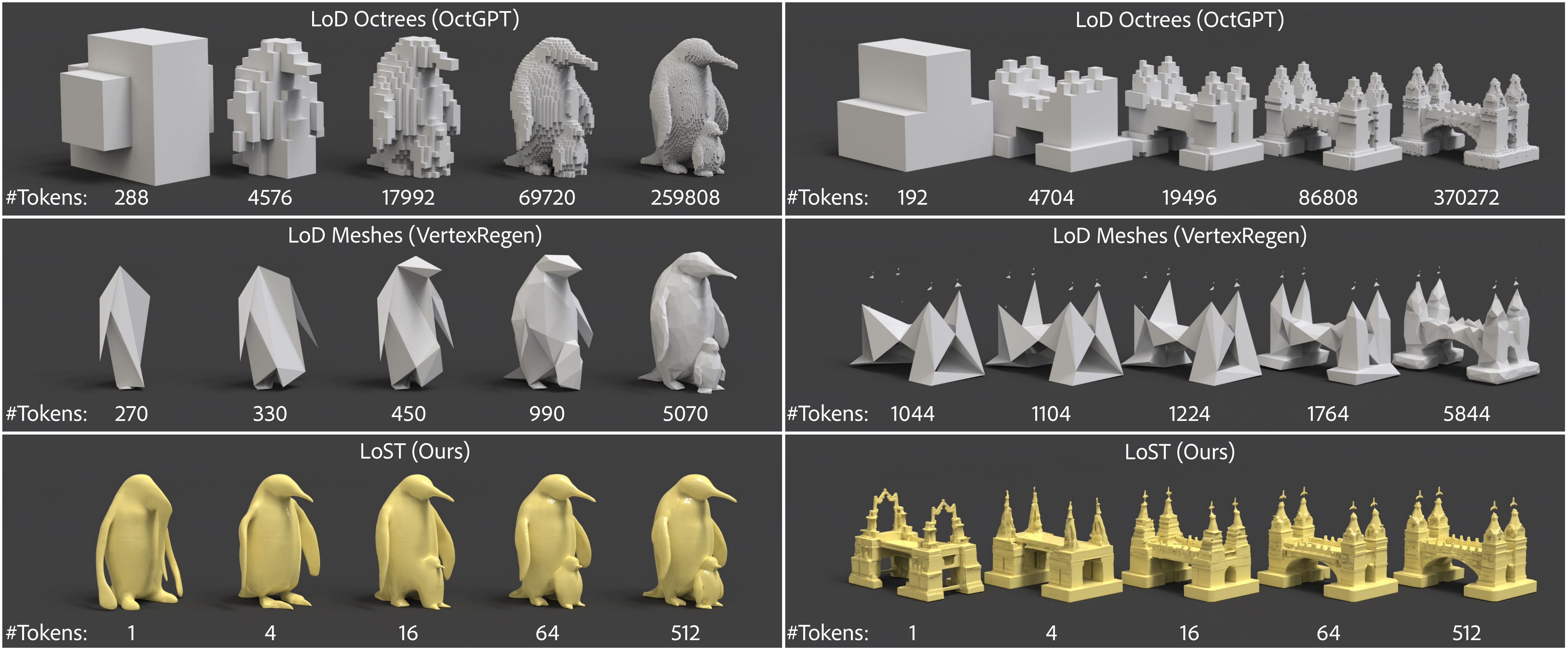}
 \captionof{figure}{
 \textbf{\name,}
a novel shape tokenization that orders tokens by semantic salience, such that early prefixes decode into complete, plausible shapes that possess principal semantics, while subsequent tokens refine instance-specific geometric and semantic details.
 \name produces prefix-decodable codes that boost semantic and geometric reconstruction over spatial level-of-detail baselines, while achieving much higher token efficiency using far fewer tokens.
 }
 \label{fig:teaser}%
 \end{center}%

 }]

\begin{abstract}

\blfootnote{\textsuperscript{*}Work done during an internship at Adobe Research.}

Tokenization is a fundamental technique in the generative modeling of various modalities. In particular, it plays a critical role in autoregressive (AR) models, which have recently emerged as a compelling option for 3D generation.
However, optimal tokenization of 3D shapes remains an open question. State-of-the-art (SOTA) methods primarily rely on geometric level-of-detail (LoD) hierarchies, originally designed for rendering and compression. These spatial hierarchies are often token-inefficient and lack semantic coherence for AR modeling.
We propose Level-of-Semantics Tokenization (\name), which orders tokens by semantic salience, such that early prefixes decode into complete, plausible shapes that possess principal semantics, while subsequent tokens refine instance-specific geometric and semantic details. 
To train \name, we introduce Relational Inter-Distance Alignment (RIDA), a novel 3D semantic alignment loss %
that aligns the relational structure of the 3D shape latent space with that of the semantic DINO feature space.
Experiments show that \name achieves SOTA reconstruction, surpassing previous LoD-based 3D shape tokenizers %
by large margins on both geometric and semantic reconstruction metrics. 
Moreover, \name achieves efficient, high-quality AR 3D generation and enables downstream tasks like semantic retrieval, while using only 0.1\%–10\% of the tokens needed by prior AR models.

\end{abstract}

 \section{Introduction}
\label{sec:intro}

Tokens have become the driving representation in generative models, spanning text, image, and video generation.
Recently, autoregressive (AR) modeling has emerged as a compelling paradigm for 3D generation. 
Compared to diffusion models, AR decoding offers simpler training, single-pass sampling, 
and seamless integration with multimodal large language models (MLLMs). 
Yet, unlike well-established tokenization in autoregressive language models, \emph{the optimal way to tokenize 3D shapes remains an open question}, despite its critical impact on the effectiveness of 3D generation and analysis.
Earlier work directly models `flat' next-element streams (voxels, points, vertices/faces~\cite{meshgpt,wang2024llamameshunifying3dmesh}), while more recent methods adopt hierarchical or multi-resolution encodings (e.g., octrees~\cite{octgpt,Wang2017_OCNN}, voxel hierarchies~\cite{Hane2020HSP}, progressive meshes~\cite{vertexregen})
to tokenize the shapes into more informative tokens guided by coarse-to-fine spatial occupancy.

We argue that such classical \emph{geometric} level-of-detail (LoD) hierarchies were originally designed for rendering and compression purposes, not for 3D shape tokenization in modern autoregressive models. 
Hence, they unfortunately suffer from several systematic issues:
(i)~\emph{token bloat} at coarse scale as even after geometric simplification, early stages still require a considerable amount of spatial tokens to sketch any object’s basic scaffold, pushing AR models into a high perplexity regime and undermining sample efficiency; 
(ii)~\emph{unusable early decoding} caused by aggressive geometric simplification used to construct geometric hierarchies, where the coarse hierarchies are overly rough and fail to resemble both geometric and semantic details of the final shape. Consequently, `any-prefix generation' produces unusable shape intermediates, limiting applicability in AR workflows.

In this work, we propose structuring shape token sequences by semantic salience, allowing short prefixes to already instantiate shapes that are \emph{plausible} and capture the original shape's %
\emph{principal semantics}, 
while subsequent tokens progressively refine the representation with instance-specific geometric and semantic details. 
To this end, we introduce \emph{Level-of-Semantics Tokenization} (\name) for  3D shapes: a learned shape token sequence \( \{\tau_1,\ldots,\tau_K\} \) in which every \emph{prefix} \( \tau_{\le k} \) decodes to a complete, plausible shape capturing principal semantics of the original shape, while longer prefixes increase instance-specific geometric and semantic details.
\Cref{fig:teaser} contrasts our level-of-semantics shape tokenization with other techniques based on level-of-detail hierarchies. For example, we can see that earlier stages in OctGPT~\cite{octgpt} and VertexRegen~\cite{vertexregen} decode into geometrically and semantically implausible shapes.

We draw inspiration from the recent Flextok~\cite{flextok} and Semanticist~\cite{semanticist} works that train an auto-encoder to learn Level of Semantics~(LoS) tokens from images.
Given a 3D shape represented by a triplane~\cite{eg3d}, we train a ViT-based shape encoder to compress the triplane features into a token sequence, while a \emph{prefix decoder} is jointly trained to reconstruct the triplane latent features from any prefix length. 
Nested token dropout and causal masking are employed to encourage 
coarse-to-fine 1D ordering of the tokens during this auto-encoder training.
Following~\cite{flextok, semanticist}, we enable the reconstruction of plausible shapes even at extreme compression rates by employing a generative decoder.

Particularly, to imbue the hierarchically ordered tokens with semantic structure, prior works~\cite{flextok, semanticist} employ an important semantic alignment loss -- REPA~\cite{repa} --
that encourages the decoder to minimize the distance between its intermediate features and the DINO features of the original image.
However, for 3D shapes, we lack the direct semantic supervision needed for this semantic alignment loss to learn level-of-semantics representations.
Hence, we introduce a \textit{3D semantics extractor} to predict semantic features of a triplane encoding using the DINO~\cite{dinov2} encoder as the teacher, inspired by Relational Knowledge Distillation (RKD)~\cite{park2019relational}.
Notably, given a triplane, the 3D semantics extractor does not directly regress DINO features obtained from its renderings. Instead, it is trained using our proposed \textit{Relational Inter-Distance Alignment (RIDA)} loss, which aligns the relative distances between samples in the triplane latent space with their corresponding semantic distances in the DINO latent space, thereby reorganizing the triplane representation according to semantic proximity in DINO space.

Evaluation demonstrates that \name sets a new state-of-the-art (SOTA) reconstruction, surpassing the previous LoD-based 3D shape tokenizer by large margins on both semantic and geometric reconstruction metrics. \name achieves this SOTA reconstruction performance while keeping a compact and semantically structured latent space suitable for autoregressive modeling. 
Autoregressive models trained on \name tokens significantly outperform SOTA models while using only 128 tokens at training and inference.
The \name tokens are also versatile and promising, extending beyond their utility in 3D autoregressive generation, as we demonstrate by showcasing their application to semantic shape retrieval.
Our contributions are summarized as follows:
\begin{itemize}
\item We introduce \name that learns to generate shape tokens ordered by semantic salience so that early prefixes can be decoded into complete and recognizable shapes capturing principal semantics, with later tokens refining instance-specific geometric and semantic details. 
\item To train \name, we design the RIDA loss, a novel 3D semantic alignment objective computed directly in triplane latent space to provide semantic supervision for learning level-of-semantics tokens for 3D shapes. 
\item We show that \name enables training a new SOTA 3D AR model with a simple GPT-style Transformer, achieving efficient, high-quality AR 3D generation, while using only 0.1\%–10\% of the tokens needed by prior 3D AR models.
\end{itemize}

 \section{Related Work}
\label{sec:related}

\paragraph{3D Tokenization with Flat Element Streams.}
Transformers that directly produce mesh elements (e.g., vertices, edges, triangles) model 3D shapes as long, irregular token streams. The seminal effort in this direction, PolyGen~\cite{polygen} autoregresses vertices and faces with a two-stage mesh model; more recently,  MeshGPT~\cite{meshgpt} and MeshXL~\cite{meshxl} treat triangles as tokens in a decoder-only transformer.
Such 1D-code streams amplify quadratic attention costs and exposure bias, and early (code) prefixes seldom decode to recognizable and/or semantically close shapes. Recently, Llama-Mesh~\cite{wang2024llamameshunifying3dmesh} unifies 3D generation and understanding with LLMs but still suffers from similar problems.

\paragraph{Learned 3D Latent Token Sequences.}
To shorten token sequences, recent works operate in compact learned 3D latent spaces~\cite{trellis, ye2025shapellm}, similar to strategies used in 2D image and video domains.
While this improves global coherence, the methods typically decode to coarse fields and rely on heavy upsamplers and/or generative diffusion for final fidelity. 
Moreover, there is no guarantee that prefixes yield complete shapes that are semantically linked. 
For instance, ShapeLLM-Omni~\cite{ye2025shapellm} mitigates some of these issues by autoregressively predicting tokens within a 3D VAE latent space, yet its generation remains limited to coarse voxel outputs, with final refinement dependent on diffusion synthesis.

\paragraph{3D Tokenization with Geometric LoD.}
Traditional hierarchical geometry (progressive meshes~\cite{Hoppe1996ProgressiveMeshes}, octrees~\cite{Samet1984QuadtreeSurvey}) yields, by construction, strong spatial coherence by emitting coarse-to-fine spatial refinements. 
Inspired by these classical representations, 
VertexRegen~\cite{vertexregen} learns vertex splits (i.e., reverse edge collapse ordering) for a more continuous LoD, while  OctGPT~\cite{octgpt} uses octrees to serialize multiscale trees for AR modeling. 
However, LoD-based encodings allocate capacity to geometric elements such as cells or edges rather than to category-defining semantics. As a result, short prefixes often decode into overly coarse shapes that lack geometric and semantic completeness.
In \Cref{sec:exp}, we compare with VertexRegen and OctGPT.

\paragraph{Hierarchical Image and Video Tokenization.}
In images and videos, discrete tokenizers and coarse-to-fine decoding have been shown to substantially improve efficiency and controllability.
VQGAN~\cite{Esser2021TamingTransformers}, as a variant of VQVAE, establishes codebook-based visual parts modeled by AR transformers; MaskGIT~\cite{Chang_2022_CVPR} introduces iterative masked decoding for rapid refinement, significantly speeding up AR decoding. Importantly, MAGVIT-v2~\cite{Yu2023MAGVITv2} shows that with a strong image/video tokenizer, AR LLMs can rival or beat diffusion on visual generation. More closely aligned to our goals, Matryoshka representation~\cite{kusupati2022matryoshka}  learns nested and prefix-usable embeddings.
More recent image tokenizers, such as FlexTok~\cite{flextok} and PCA-like Semanticist~\cite{semanticist}) for images, explicitly order tokens by semantic salience, enabling variable-length token outputs.
Inspired by these, we seek a 3D tokenizer that ensures an \emph{any-prefix} decoder that is both semantically relevant and geometrically refined.

 \section{Method}
\label{sec:method}

\newcommand{\register}{\ensuremath{ \mathcal{T}_R }\xspace}
\newcommand{\shapetoken}{\ensuremath{ \mathcal{T}_{3D} }\xspace}
\newcommand{\diff}{\ensuremath{\mathcal{G}}\xspace}
\newcommand{\noisy}{\ensuremath{x_t}\xspace}
\newcommand{\clean}{\ensuremath{x_0}\xspace}
\newcommand{\noise}{\ensuremath{\epsilon}\xspace}

\begin{figure*}[tp]
    \centering
    \includegraphics[width=\linewidth]{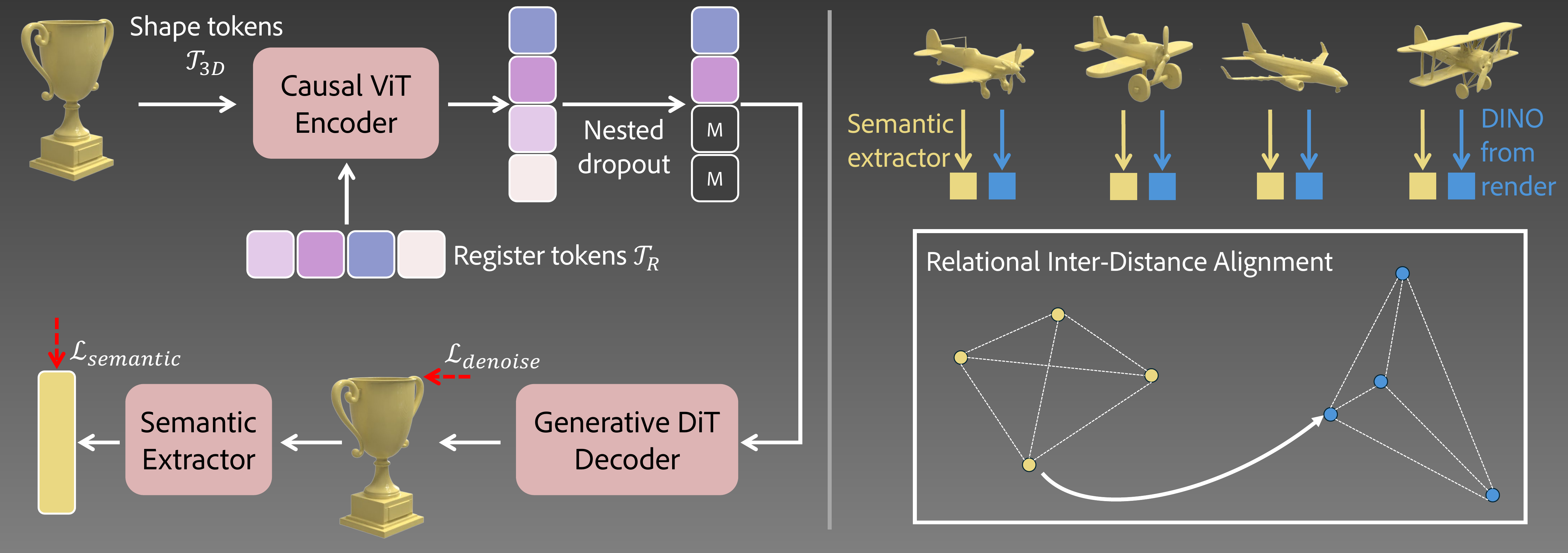}
    \caption{\textbf{Overview of LoST}. Left: LoST maps 3D shape latents into a token sequence ordered by semantic salience, where early prefixes capture coarse semantics and later tokens refine instance-specific detail. A conditional generative DiT decoder reconstructs the complete latent from any prefix. Right: The semantic extractor is pretrained with Relational Inter-Distance Alignment (RIDA), which aligns relationships in 3D latent space with DINO feature relationships to provide semantics-aware supervision.}
    \label{fig:overview}
\end{figure*}

Our goal is to learn token sequences structured by semantic salience with the following properties: (i) earlier prefixes already instantiate shapes that are \emph{plausible} and capture the original shapes' %
\emph{principal semantics}, 
(ii) subsequent tokens progressively refine the representation with instance-specific geometric and semantic details. 
In the following, we present the proposed \emph{Level-of-Semantics Tokenization} (\name) in detail and describe the key algorithmic components that enable its effective training.
\Cref{fig:overview} presents an overview.

\subsection{\name Encoder}

Following common practice in the field, we start from VAE-encoded 3D shapes, which provide a smooth and compact latent space.
In our work, we adopt the VAE learned in Direct3D~\cite{direct3d}, which encodes a shape’s point cloud into a triplane of size $\mathbb{R}^{C \times H \times W \times 3}$, yielding $32 \times 32 \times 3 = 3072$ feature vectors, each of $C=16$ dimensions. To transform the triplane into a 1D token sequence $\shapetoken$,
we employ a ViT~\cite{vit}-based encoder on patchified triplanes, following common practice~\cite{tiktok}.
However, as
each of these tokens is associated with a triplane patch,  restructuring their content to represent semantic LoDs is difficult.
Instead, we introduce a new set of \emph{register}~\cite{darcet2023vision} tokens, \register, designed to capture this hierarchical semantic signal.
These register tokens are learnable parameters that are concatenated with the triplane tokens and processed through the attention layers of the ViT. Unlike the original tokens $\shapetoken$, they are not associated with a triplane patch and can be used to hold a summarized representation of the original tokens.
The attention is masked so that the register tokens can attend to the original tokens, but not vice versa.
After transformer encoding, only the register tokens are retained, while the original tokens are discarded. 
This effectively restructures the geometric information from the triplane tokens into a learned 1D token sequence \register.

To ensure \register forms a \emph{hierarchical} token sequence, we adopt several strategies following~\cite{flextok, semanticist, nested}:
(i) we apply causal masking for \register in the ViT encoder to encourage a hierarchical structure; and importantly, (ii) we use nested dropout~\cite{nested} to enforce earlier tokens to capture the principal semantics of the representation, while the subsequent tokens add finer details. 
During training, only a prefix of \register with random length is kept while masking out the remainder (see \Cref{fig:overview}, top). 
In practice, we sample prefix lengths that are powers of 2, i.e., $[1,2,4,8,... k]$. This naturally forces the model to front-load coarse
information into the first few tokens, while later tokens progressively encode
finer
details, resulting in a hierarchical structure. The type of hierarchy depends on the type of loss used to train the encoder: a geometric loss gives us a geometric hierarchy of low- to high-frequency details, 
analogous to spectral analysis in 3D geometry~\cite{Vallet2008ManifoldHarmonics}, and a semantic loss gives us a more semantic hierarchy.
We use 768 triplane tokens after patchification and a maximum of $k=512$ register tokens \register as we see marginal improvement beyond this.

\subsection{\name Decoder}
We aim to decode the full sequence of 3D latents from any prefix of the register tokens. However, reconstructing the complete geometric signal from very few tokens is inherently challenging,
as the ambiguity inherent in the limited information results in blurry, coarse reconstructions when decoded deterministically.
Instead of exact geometric reconstruction from very few tokens, 
we focus on producing \emph{semantically plausible} reconstructions that may differ in geometry. 
To this end, following~\cite{flextok, semanticist}, we reframe the task as a generative problem and employ a diffusion model to produce the full sequence conditioned on a variable-length prefix of the encoded register tokens. As the prefix length increases, generation gradually transitions toward reconstruction, since longer prefixes reduce ambiguity in the predicted sequence.
More concretely,
we train a Diffusion-Transformer (DiT) model~\cite{ddpm, scalableDiT}, \diff to reproduce the full signal conditioned on a flexible prefix of \register, which is obtained by simply masking out the unused postfix.
The generator \diff takes as input noisy shape tokens  and predicts the added noise by cross-attending to the conditional \register. See supplemental for details.

\subsection{Semantic Guidance for Learning LoS Tokens}
To improve the semantic structure,
both FlexTok and Semanticist \cite{flextok, semanticist} have relied on Representation Alignment~\cite{repa} (REPA) loss to enforce alignment between internal representations of the diffusion model and semantic DINO~\cite{dinov2} features extracted from the target image. 
The inclusion of such a DINO-based semantic REPA loss
encourages the tokens to encode semantics and 
enables the learned hierarchy to capture progressively richer levels of semantics within the token sequence.
However, no comparable semantic feature extractor and alignment loss exist for 3D shape generation, making it challenging to directly apply REPA-style supervision in our setting.
Directly aligning the internal representations of our 3D generative model with those of a 2D visual foundation model (\emph{e.g.}, DINO) performs poorly, even when reconstructing from the complete set of register tokens.
The failure can be attributed to
differences in the spatial layout and inherent dimensionality of the two representations.
An option to align the dimensionality of the two representations could be to apply the REPA loss to multi-view renders of decoded triplanes, but this is computationally prohibitive.

\paragraph{Relational Inter-Distance Alignment (RIDA).}
Our key insight is that we only
need to align
\emph{contrastive relative distances} 
in corresponding sample sets of the two representations, rather than regressing absolute values.
Therefore, we define a mapping from the triplane latent space into a new feature space where relative distances match those of DINO. Once this mapping is established, we can use this feature space instead of DINO for semantic guidance.
Specifically, we propose \emph{Relational Inter-Distance Alignment (RIDA)}, a novel pre-training for creating this mapping 
to
a \emph{student} feature space, where relative distances are aligned to a \emph{teacher}.

In our setting, the teacher space is formed by DINO features. As our training set consists of 3D shapes reconstructed from generated images, we encode DINO features directly from the generated images, giving us spatial tokens \(\mathbf{S}_i^t\in\mathbb{R}^{K\times d}\) and a further global embedding \(\mathbf{z}^{t}_i\in\mathbb{R}^{D}\).
To obtain the student space, 
we train a transformer-based encoder \(f_\theta\) that maps a triplane encoding \(\mathbf{X}_i\) to the new student space. Analogous to the teacher space, the student space consists of a semantic spatial grid \(\mathbf{S}_i^s\in\mathbb{R}^{K\times d}\) and a further global embedding \(\mathbf{z}^{\,s}_i\in\mathbb{R}^{D}\) obtained by attention pooling over the grid. We call $f_\theta$ the \emph{semantic extractor}.

The semantic extractor is trained to ensure that the relational topology of the student space mimics that of the teacher space. Note that the features themselves are not directly comparable between the two spaces, as they encode different modalities (images vs. 3D shapes).
To learn contrastive semantic relationships, the teacher space is used to mine a \emph{positive set} \(\mathcal{Z}_i^+\subset\{\mathbf{z}_{i,1}^{s,+},\dots,\mathbf{z}_{i,m}^{s,+}\}\) and a \emph{negative set} \(\mathcal{Z}_i^-\subset\{\mathbf{z}_{i,1}^{s,-},\dots,\mathbf{z}_{i,m}^{s,-}\}\) for each anchor $\mathbf{X}_i$ based on specified thresholds. This teacher-guided mining dictates which pairs should be pulled together and which should be pushed apart in the student space. 
Below, we describe the objectives used to train our 3D semantic extractor $f_\theta(\mathbf{X}_i)$.

\paragraph{Global Relational Contrast.}
First, we use the mined positive $\mathcal{Z}_i^+$ and negative $\mathcal{Z}_i^-$ sets to structure the global embedding space. We adopt a multi-positive InfoNCE loss~\cite{infonce} that pulls the student anchor $\mathbf{z}^{\,s}_i$ towards all of its teacher-defined positives $\mathcal{Z}_i^+$, while pushing it away from the negatives $\mathcal{Z}_i^-$. 
Let $\mathcal{B}\subset\{\mathbf{z}_1^s,\dots,\mathbf{z}_p^s\}$ be the set of all embeddings in the current training batch of size $p$, and $c_{ij} = \langle \mathbf{z}_i, \mathbf{z}_j \rangle$ be the cosine similarity, we define: 
\begin{equation}
\label{eq:global_contrast}
\mathcal{L}_{\mathrm{global}}
:= -\mathbb{E}_{\mathbf{z}_i \in \mathcal{B}}
\left[
\log
\frac{\sum_{\mathbf{z}_j\in\mathcal{Z}_i^+}\exp(c_{ij})}
     {\sum_{\mathbf{z}_k\in(\mathcal{Z}_i^+\cup\mathcal{Z}_i^-)}\exp(c_{ik})}
\right].
\end{equation}
This loss
ensures that semantically similar 3D shapes are mapped to nearby points in the student's latent space.

\paragraph{Inter-Instance Rank Distillation.}
The contrastive loss enforces separation based on hard thresholds between positive and negative samples, but discards the rich, continuous relational structure within the teacher's space. 
This continuous structure is essential, but it is non-trivial to transfer to the student space. 
To this end, we are inspired by Relational Knowledge Distillation (RKD)~\cite{park2019relational}, which transfers pairwise Euclidean distances, and introduce the inter-instance rank distillation loss $\mathcal{L}_{\mathrm{rank}}$ for additional supervision.

\paragraph{Spatial Structure Distillation.}
To ensure the student's spatial tokens capture the same part-level relationships as the teacher's, we distill the intra-instance token affinities, and introduce the spatial structure distillation loss $\mathcal{L}_{\mathrm{spatial}}$ as an additional training objective.

The final semantic pretraining objective for our student encoder \(f_\theta\) is now a weighted sum of these components:
\begin{equation}
\label{eq:total_sem_loss}
\mathcal{L}_{\mathrm{RIDA}}
:= \lambda_{\mathrm{g}}\mathcal{L}_{\mathrm{global}}
+ \lambda_{\mathrm{r}}\mathcal{L}_{\mathrm{rank}}
+ \lambda_{\mathrm{s}}\mathcal{L}_{\mathrm{spatial}}.
\end{equation}
We use $\lambda_{\mathrm{global}}=1.0$, $\lambda_{\mathrm{rank}}=1.0$, and $\lambda_{\mathrm{spatial}}=0.5$ in our experiments. The resulting network \(f_\theta\) provides a semantically-structured 3D latent space, with which we can now guide the LoST learning.
Details of $\mathcal{L}_{\mathrm{rank}}$ and $\mathcal{L}_{\mathrm{spatial}}$ are presented in the supplementary.

\paragraph{Semantic-guided \name Training.}
With the semantic encoder $f_\theta$ pre-trained using RIDA, we employ it as a perceptual loss to guide the diffusion generator $\mathcal{G}$. This semantic alignment loss, $\mathcal{L}_{\text{semantic}}$, maximizes the cosine similarity between $\mathcal{G}$'s predicted latent $\hat{\mathbf{X}}_0$ and the ground-truth latent $\mathbf{X}_0$. Specifically, 
\begin{equation}
\label{eq:sem_gen_loss}
\mathcal{L}_{\text{semantic}} := \mathbb{E}_{t, \mathbf{X}_0, \epsilon} \left[ 1 - \langle f_\theta(\hat{\mathbf{X}}_0), f_\theta(\mathbf{X}_0) \rangle \right].
\end{equation}
The final objective for training the generator $\mathcal{G}$ combines the geometric fidelity loss $\mathcal{L}_{\text{denoise}}$ with our semantic loss:
\begin{equation}
\label{eq:final_loss}
\mathcal{L} := \mathcal{L}_{\text{denoise}} + \lambda_{\text{semantic}} \mathcal{L}_{\text{semantic}}.
\end{equation} 
We use $\lambda_{\text{semantic}}=1.0$ in our experiments.

\subsection{\name-GPT}
Differing from prior work on 3D autoregressive generation, we do not quantize the tokenizer outputs. Instead, we keep our \register in continuous space. We then train a GPT-style Transformer, following the standard setup of LlamaGen~\cite{llamagen}, to autoregressively model these continuous \register tokens. Rather than using a categorical cross-entropy loss, we adopt a diffusion loss~\cite{ddpm} following MAR~\cite{ar_diff_loss}, which shows that autoregressive models can perform next-token prediction in continuous space by modeling the per-token conditional distribution with a small MLP. Concretely, at each position the Transformer predicts a conditioning vector, and a small MLP-based diffusion head, conditioned on this vector, maps this to the final token. For conditional generation, we utilize OpenCLIP~\cite{openclip,Radford2021LearningTV} embeddings, which are prepended to the input sequence so that the conditioning information is propagated throughout the next-token prediction process.

 \section{Experiments}
\label{sec:exp}

\begin{figure*}[t]
    \centering
    \includegraphics[width=\linewidth]{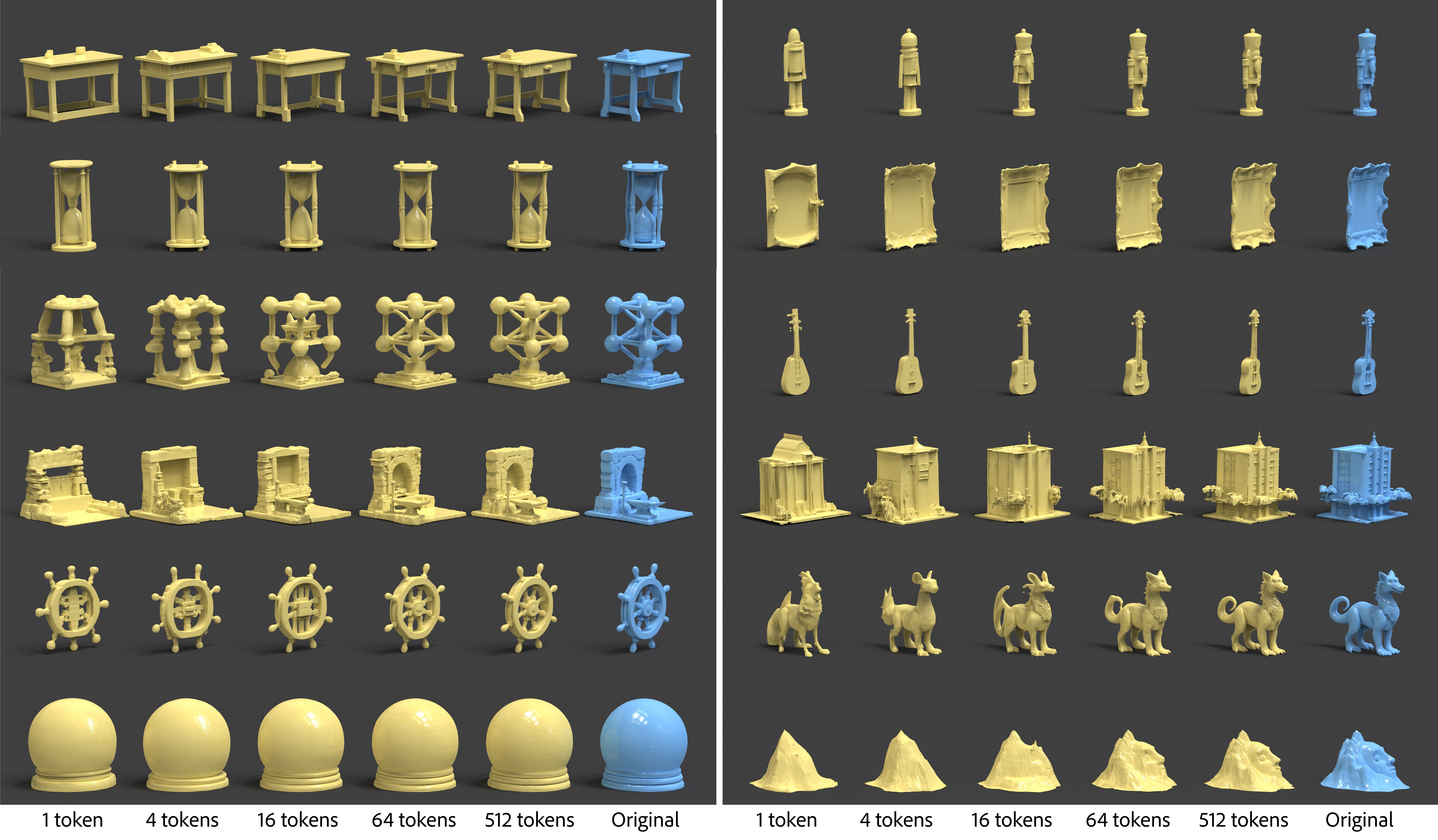}
    \caption{
    For each 3D shape (in blue), we visualize the shapes (in yellow) decoded from the learned \name token sequences. Even as few as 1 token based generations result in semantically similar shapes while more tokens help to capture both semantic and geometric details.
    }
    \label{fig:tokenization_gallery}
\end{figure*}

\label{subsec:implementation}
\paragraph{Training Dataset.} 
\name is trained on the latent space of Direct3D's VAE~\cite{direct3d}. 
Rather than relying on the large-scale Objaverse dataset~\cite{deitke2023objaverse}, which requires substantial preprocessing,
we opted to generate our own training dataset for minimum overhead and maximum compatibility with Direct3D. This was done by directly rolling out generation samples from Direct3D via its image-to-3D pipeline.
To create the dataset, we first generated diverse prompts for varied objects 
using Gemini 2.5 Pro~\cite{comanici2025gemini25pushingfrontier}. These prompts were then used for image synthesis via Flux.1~\cite{flux2024}, and the resulting images were subsequently lifted to 3D shapes. This yielded a dataset of 300k shapes. 
Please refer to our supplementary for the prompt template.

\noindent\textbf{Tokenizer.}
Our ViT encoder is configured with a depth of 12, an embedding dimension of 768, and 16 attention heads. Following encoding, the registers are projected to a dimension of 32 (\register $\in \mathbb{R}^{32}$), which serves as the information bottleneck. We train a DiT decoder with a depth of 24, a hidden dimension of 1024, and 16 heads. Both models utilize $2 \times 2$ patchification. The \name model is trained for 250 epochs using 8xA100 GPUs.  We employ a random dropout rate of 0.1 for classifier-free guidance and 2D sinusoidal positional embeddings.

\noindent\textbf{RIDA.}
The RIDA model is a student Transformer configured with a depth of 12, an embedding dimension of 768, and 8 attention heads. It is trained for 100 epochs to distill features from a frozen DINOv2 ViTB14 teacher~\cite{dinov2}. The model's tokenizer first processes the 3D triplane latents by splitting them into three planes, preserving the 3D structure. Each plane is passed through a stem of 2D depth-wise convolutions before being fused. A final strided convolution acts as a patchification step, producing 256 tokens ($\mathbb{R}^{16 \times 16 \times 768}$). These tokens, augmented with 2D sinusoidal positional embeddings, are fed into the Transformer.

\noindent\textbf{AR Generation.}
Our LlamaGen based AR model has attention heads of depth 24, 16 attention heads, and a hidden dimension of 1024. 
We employ a random dropout rate of 0.1 for classifier-free guidance. We train \name-GPT on 128 tokens to balance efficiency and fidelity.

\begin{table*}[t]
    \centering %
    \caption{\textbf{Tokenizer reconstruction.} We compare the reconstruction of \name to recent LoD based tokenizers (OctGPT and VertexRegen) using varying number of tokens. We report the Chamfer Distance (CD) for geometric, and FID and DINO similarity for semantic accuracy. \name outperforms baselines while using significantly fewer tokens. The top-performing score for each decoding level is in \textbf{bold}.}
    \label{tab:tokenizer}
    
    \resizebox{\textwidth}{!}{
        \begin{tabular}{r | rrrrr | rrrrr | rrrrr}
            \toprule
            
            {} 
            & \multicolumn{5}{c}{OctGPT~\cite{octgpt}} 
            & \multicolumn{5}{c}{VertexRegen~\cite{vertexregen}} 
            & \multicolumn{5}{c}{\textbf{LoST (ours)}} \\
            
            \cmidrule(lr){2-6} \cmidrule(lr){7-11} \cmidrule(lr){12-16}
            
            Num Tokens $\rightarrow$ & \textbf{$\sim$219} & \textbf{$\sim$3615} & \textbf{$\sim$15031} & \textbf{$\sim$61962} & \textbf{$\sim$239004} 
            & \textbf{$\sim$2730} & \textbf{$\sim$2790} & \textbf{$\sim$2910} & \textbf{$\sim$3450} & \textbf{$\sim$7530}
            & \textbf{1} & \textbf{4} & \textbf{16} & \textbf{64} & \textbf{512} \\
            
            \midrule

            CD ($\times 10^{-2}$)$\downarrow$
            & 16.923 & 1.759 & 0.850 & 0.533 & 0.470
            & 4.290 & 1.865 & 0.809 & \textbf{0.209} & \textbf{0.034}
            & \textbf{2.271} & \textbf{1.328} & \textbf{0.723} & 0.382 & 0.234 \\
            
            FID$\downarrow$
            & 341.174 & 265.774 & 184.252 & 100.781 & 88.483
            & 186.611 & 186.454 & 176.137 & 151.393 & 86.098
            & \textbf{31.649} & \textbf{29.255} & \textbf{26.565} & \textbf{21.133} & \textbf{13.591} \\
            
            DINO$\uparrow$
            & 0.382 & 0.470 & 0.535 & 0.619 & 0.695
            & 0.463 & 0.485 & 0.518 & 0.602 & 0.791
            & \textbf{0.731} & \textbf{0.765} &\textbf{ 0.814} & \textbf{0.880} & \textbf{0.921} \\
            
            \bottomrule
        \end{tabular}
    } %

\end{table*}

\subsection{Tokenization Evaluation}
\label{eval:tokenizer}

\paragraph{Baselines.} 
We compare \name against recent Level-of-Detail (LoD) tokenizers operating at various decoding granularities: (i) OctGPT~\cite{octgpt} utilizes OctTrees for hierarchical representations, and (ii) VertexRegen~\cite{vertexregen} is based on an iterative edge-collapse strategy. We use their recommended token hierarchy levels for comparison across token levels.

\noindent\textbf{Evaluation Dataset.}
For robust reconstruction evaluation, we curate a novel, unseen test set of 1k shapes. 
Compared to the relatively simple, clean CAD-style objects in ShapeNet~\cite{chang2015shapenet} and Toys4K, our test shapes exhibit more complex geometry. 
We follow the same text-prompt protocol as our training data, but instead synthesize this set using Step1X-3D~\cite{li2025step1x} image-to-3D pipeline. 
Importantly, Step1X-3D is built on the 3DShape2VecSet~\cite{shape2vecset} representation, whereas LoST is trained on triplane latents. 
This makes the evaluation \emph{neutral}: the test shapes are produced with an unseen set of samples generated from new prompts, by an independent SOTA 3D generative model with a different internal representation and architecture than ours. The resulting meshes undergo post-processing to ensure clean, watertight geometry, including the removal of degenerate faces and face count reduction. We also perform necessary mesh processing as required for each baseline method.

\begin{figure}[t]
    \centering
    \includegraphics[width=\linewidth]{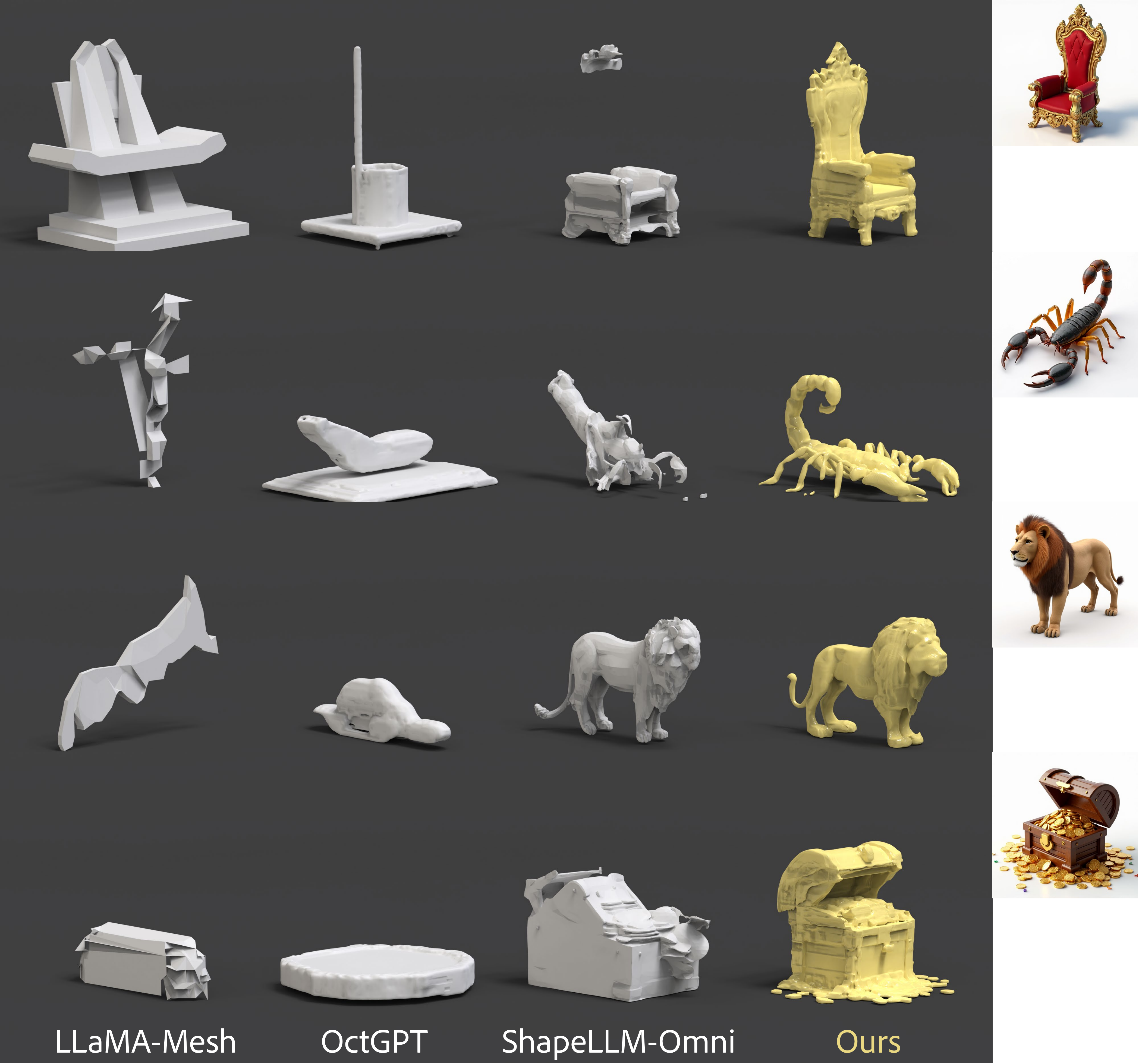}
    \caption{\textbf{AR generation comparison.} We compare image/text based autoregressive generation methods. \name achieves superior performance in high-quality and faithful generation.}
    \label{fig:ar_comparison}
\end{figure}

\noindent\textbf{Metrics}.
Our evaluation utilizes two complementary metrics to assess both geometric and semantic fidelity.
Geometric accuracy is quantified using the Chamfer Distance (CD), while semantic consistency is evaluated by computing the DINO cosine similarity and FID~\cite{fid} of 2D renderings of the reconstructed against ground-truth target shapes.

\paragraph{Discussion.}
 We report quantitative results in \Cref{tab:tokenizer}. \name significantly outperforms baselines on semantic (DINO, FID) and geometric (CD) metrics, especially at low token budgets. The ability to generate plausible, complete shapes from short prefixes—unlike the abstract scaffolds of LoD methods (\Cref{fig:teaser}) results in superior early fidelity. \name achieves better reconstruction and alignment using just 0.1\%-10\% of the tokens required by baselines; even using 1-4 tokens often surpass them. As token count increases, generation variance decreases as the task shifts from plausible generation to high-fidelity reconstruction.

Qualitative results (\Cref{fig:tokenization_gallery}, \Cref{fig:teaser}) confirm this Level-of-Semantics progression. A single \name token decodes a complete, recognizable shape, with later tokens refining instance-specific details, such as progressing from a generic mountain to one with an embedded face. Most importantly, we see semantically plausible high frequency details at all levels. In contrast, LoD methods produce implausible primitives. For simple shapes, like the crystal ball in \Cref{fig:tokenization_gallery}, 1-4 tokens is often sufficient. Please refer to the supplementary for additional results, including ablations.

\begin{figure}[t!]
    \centering
    \includegraphics[width=1.\linewidth]{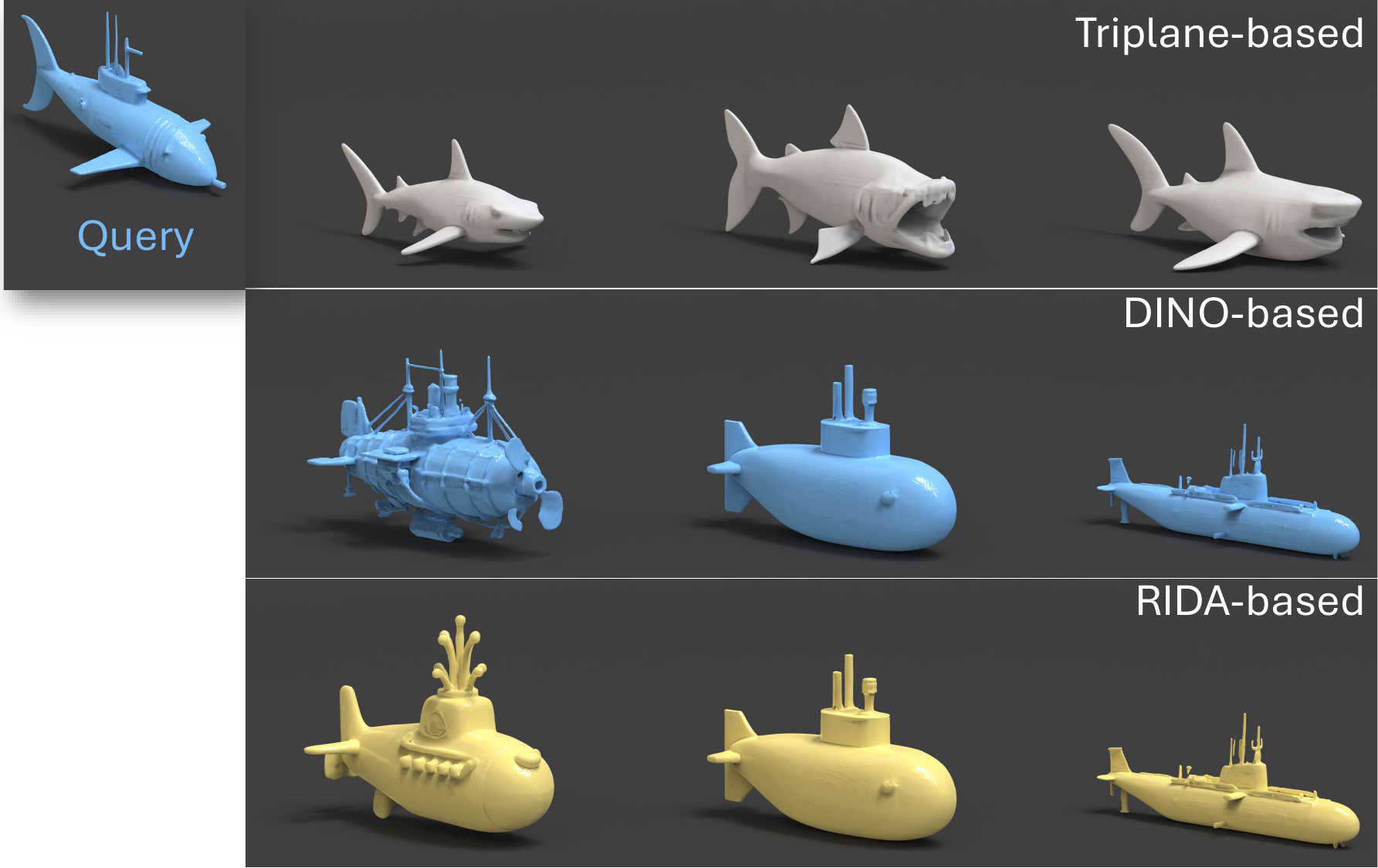}
    \caption{Given a query shape (top), we show shape retrieval results using triplane, DINO, and RIDA features. While original triplane features focus on geometric similarity, RIDA mapped triplane features capture semantic alignment similar to DINO. In this example, we use a confusing query of a submarine shaped like a fish.}
    \label{fig:retrieval}
    \vspace{-10pt}
\end{figure}

\begin{figure}[t]
    \centering
    \includegraphics[width=\linewidth]{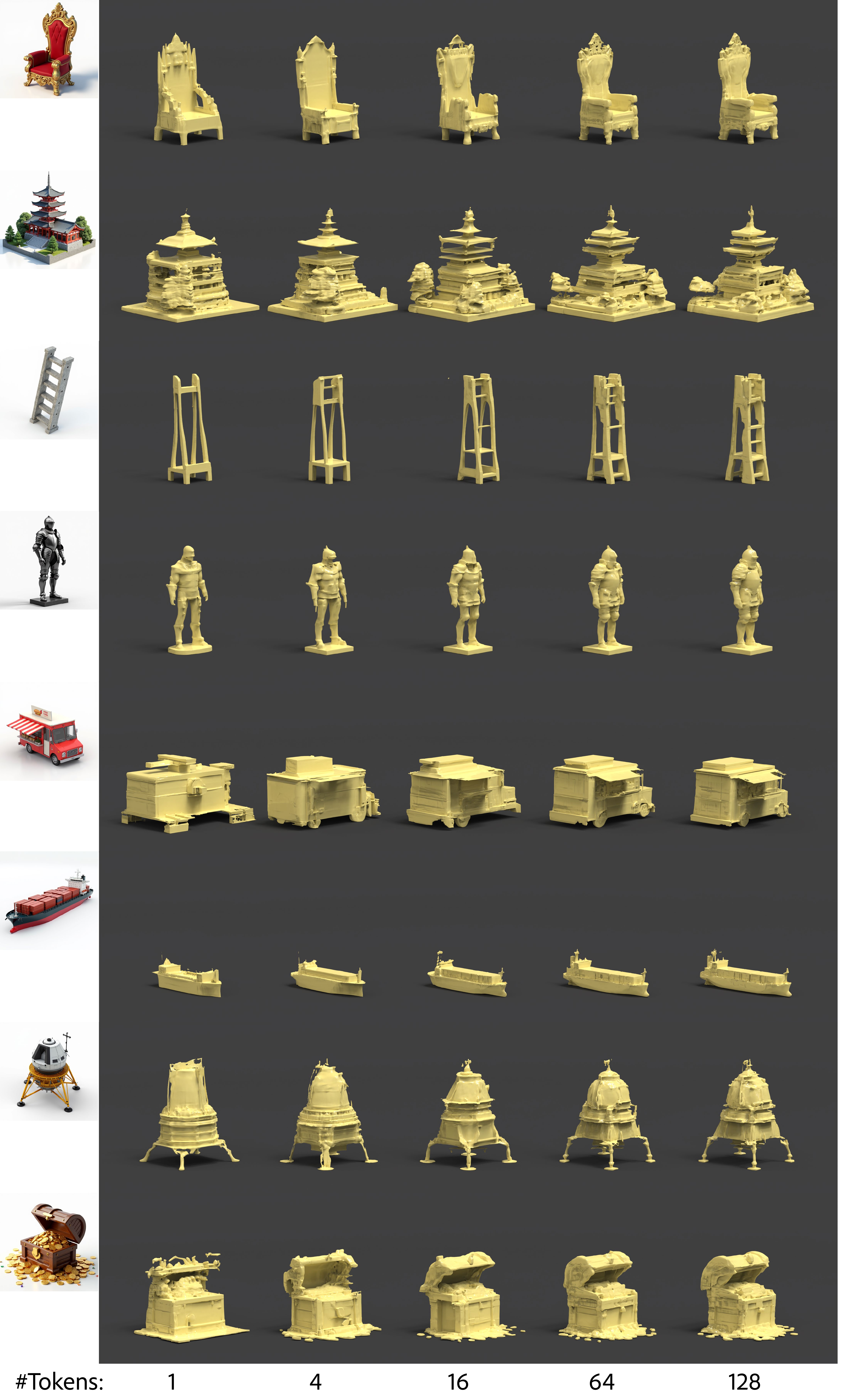}
    \caption{\name-GPT trained with our tokenizer can generate varying number of token sequences that can be decoded into complete, level-of-semantics shapes enabling efficient early stopping. 
    For example, the model predicts a treasure chest without coins and a cargo ship without containers at the 1-token level, with these finer semantic and geometric details added in later levels.
    }
    \label{fig:ar_multi_level}
\end{figure}

\subsection{Autoregressive Generation Evaluation}
\label{eval:autoregressive}

\paragraph{Baselines and Setup.} 
We evaluate the generative capabilities of \name by comparing against existing SOTA 3D AR generation models: ShapeLLM-Omni~\cite{ye2025shapellm}, OctGPT~\cite{octgpt}, and Llama-Mesh~\cite{wang2024llamameshunifying3dmesh}. All methods are trained on the large-scale 3D repository, Objaverse~\cite{deitke2023objaverse}.
We omit VertexRegen from this analysis, as the public codebase is restricted to tokenization and does not include AR generation. OctGPT and Llama-Mesh support \textbf{Text-to-3D} synthesis, while our model and ShapeLLM-Omni are evaluated on the \textbf{Image-to-3D} task (we use the same text prompts for generating images with Flux). 
We consider the same set of text prompts and synthesized 3D shapes as in \Cref{eval:tokenizer} to ensure consistency across tokenization and generation evaluation.

\begin{table}[t]
  \centering
  \footnotesize   
  \caption{\textbf{AR generation.} We report shading-based FID along with DINO similarity for image-conditioned methods. We use renderings of the generated evaluation dataset as ground truth reference.}
  \label{tab:main_results}
  \begin{tabular}{r|ccc}
    \toprule
    Method & Num Tokens & FID $\downarrow$ & DINO $\uparrow$ \\
    \midrule
    OctGPT~\cite{octgpt}   & $\sim$50,000       & 66.926                & \ding{55} \\
    Llama-Mesh~\cite{wang2024llamameshunifying3dmesh}  & $\sim$3758     & 118.576               & \ding{55} \\
    ShapeLLM-Omni~\cite{ye2025shapellm} & 1024  & 48.702              & 0.680 \\
    \name (ours) & 128 & \textbf{34.251}               & \textbf{0.758} \\
    \bottomrule
  \end{tabular}
\end{table}

\noindent\textbf{Metrics.}
We compute the FID score to measure the distributional alignment between the generated shape renderings and the target shape renderings.
For models evaluated on image-to-3D (ShapeLLM-Omni and \name), we additionally report the DINO cosine similarity between the generated and target shape renderings, following \Cref{eval:tokenizer}.

\noindent\textbf{Discussion.}
We compare \name-GPT against SOTA baselines in \Cref{tab:main_results}. Quantitatively, \name-GPT significantly outperforms all competitors, achieving the lowest FID and the highest DINO semantic alignment score on image-to-3D generation, setting a SOTA for autoregressive 3D generation. This is achieved while training on only 128 tokens, underscoring the efficiency of our \name tokens. We note that ShapeLLM-Omni~\cite{ye2025shapellm} is a two-stage method where an AR model predicts coarse voxels, and a refiner (Trellis~\cite{trellis}) generates the final geometry. Despite this refinement stage, \name-GPT achieves superior performance on both metrics.
OctGPT and Llama-Mesh are text-to-3D generation methods, therefore, we evaluate them using the text prompts used to generate our evaluation dataset. Consequently, we cannot compute the DINO score, which measures alignment with a conditional image , and thus omit it for these models in \Cref{tab:main_results}. Qualitative comparisons in \Cref{fig:ar_comparison} further highlight the gap between ours and baselines. \name-GPT consistently generates high-fidelity, semantically coherent shapes, while baselines often produce abstract, incomplete, or malformed results. Moreover, the autoregressive nature combined with our ``any-prefix" tokenization allows for efficient inference, as \name-GPT can be early-stopped when generating simpler shapes that are fully captured with fewer tokens. We show AR results at varying token lengths in \Cref{fig:ar_multi_level}.

 \section{Conclusion}

We revisited tokenization for 3D AR generation and argued that geometric level-of-detail is a poor organizing principle for next-token prediction. 
Instead, we introduce \emph{Level-of-Semantics Tokenization} (\name), which orders tokens by semantic salience so that short prefixes already decode complete, category-plausible shapes.
For training, we proposed \emph{Relational Inter-Distance Alignment} (RIDA), a 3D semantic loss that couples triplane latents to an image-semantic space (DINO) without the need for the computationally expensive decoding-and-rendering process.

Despite the demonstrated success, we note a few shortcomings: 
Our tokenizer and losses are instantiated on VAE triplane latents. Extending \name to
support other 3D representations such as Gaussian Splats would be a natural next step;
We use a diffusion decoder to produce the final latents from the AR generated tokens, which increases computational requirements compared to pure AR decoding;
Although \name improves early-prefix usability, the few-token regime can still exhibit artifacts compared to full-length decodes - also observed in 2D semantics-first tokenizers. Future work includes strengthening early tokens with topology-aware regularizers and part-consistency constraints;
Finally, although \name produces variable-length codes, our AR generator currently uses a fixed target length. Adding an \texttt{EOS} token and complexity-aware stopping (shorter for simple shapes, longer for complex ones) is a natural extension.

 {
     \small
     \bibliographystyle{ieeenat_fullname}
     \bibliography{references}

\begin{thebibliography}{43}
\providecommand{\natexlab}[1]{#1}
\providecommand{\url}[1]{\texttt{#1}}
\expandafter\ifx\csname urlstyle\endcsname\relax
  \providecommand{\doi}[1]{doi: #1}\else
  \providecommand{\doi}{doi: \begingroup \urlstyle{rm}\Url}\fi

\bibitem[Bachmann et~al.(2025)Bachmann, Allardice, Mizrahi, Fini, Kar, Amirloo, El-Nouby, Zamir, and Dehghan]{flextok}
Roman Bachmann, Jesse Allardice, David Mizrahi, Enrico Fini, O{\u{g}}uzhan~Fatih Kar, Elmira Amirloo, Alaaeldin El-Nouby, Amir Zamir, and Afshin Dehghan.
\newblock Flextok: Resampling images into 1d token sequences of flexible length.
\newblock In \emph{Forty-second International Conference on Machine Learning}, 2025.

\bibitem[Chan et~al.(2022)Chan, Lin, Chan, Nagano, Pan, De~Mello, Gallo, Guibas, Tremblay, Khamis, et~al.]{eg3d}
Eric~R Chan, Connor~Z Lin, Matthew~A Chan, Koki Nagano, Boxiao Pan, Shalini De~Mello, Orazio Gallo, Leonidas~J Guibas, Jonathan Tremblay, Sameh Khamis, et~al.
\newblock Efficient geometry-aware 3d generative adversarial networks.
\newblock In \emph{Proceedings of the IEEE/CVF conference on computer vision and pattern recognition}, pages 16123--16133, 2022.

\bibitem[Chang et~al.(2015)Chang, Funkhouser, Guibas, Hanrahan, Huang, Li, Savarese, Savva, Song, Su, et~al.]{chang2015shapenet}
Angel~X Chang, Thomas Funkhouser, Leonidas Guibas, Pat Hanrahan, Qixing Huang, Zimo Li, Silvio Savarese, Manolis Savva, Shuran Song, Hao Su, et~al.
\newblock Shapenet: An information-rich 3d model repository.
\newblock \emph{arXiv preprint arXiv:1512.03012}, 2015.

\bibitem[Chang et~al.(2022)Chang, Zhang, Jiang, Liu, and Freeman]{Chang_2022_CVPR}
Huiwen Chang, Han Zhang, Lu Jiang, Ce Liu, and William~T. Freeman.
\newblock Maskgit: Masked generative image transformer.
\newblock In \emph{CVPR}, pages 11315--11325, 2022.

\bibitem[Chen et~al.(2024)Chen, Chen, Pang, Zeng, Cheng, Fu, Yin, Wang, Yu, Yu, et~al.]{meshxl}
Sijin Chen, Xin Chen, Anqi Pang, Xianfang Zeng, Wei Cheng, Yijun Fu, Fukun Yin, Billzb Wang, Jingyi Yu, Gang Yu, et~al.
\newblock Meshxl: Neural coordinate field for generative 3d foundation models.
\newblock \emph{Advances in Neural Information Processing Systems}, 37:\penalty0 97141--97166, 2024.

\bibitem[Darcet et~al.(2023)Darcet, Oquab, Mairal, and Bojanowski]{darcet2023vision}
Timoth{\'e}e Darcet, Maxime Oquab, Julien Mairal, and Piotr Bojanowski.
\newblock Vision transformers need registers.
\newblock \emph{arXiv preprint arXiv:2309.16588}, 2023.

\bibitem[Deitke et~al.(2023)Deitke, Schwenk, Salvador, Weihs, Michel, VanderBilt, Schmidt, Ehsani, Kembhavi, and Farhadi]{deitke2023objaverse}
Matt Deitke, Dustin Schwenk, Jordi Salvador, Luca Weihs, Oscar Michel, Eli VanderBilt, Ludwig Schmidt, Kiana Ehsani, Aniruddha Kembhavi, and Ali Farhadi.
\newblock Objaverse: A universe of annotated 3d objects.
\newblock In \emph{Proceedings of the IEEE/CVF conference on computer vision and pattern recognition}, pages 13142--13153, 2023.

\bibitem[Dosovitskiy et~al.(2021)Dosovitskiy, Beyer, Kolesnikov, Weissenborn, Zhai, Unterthiner, Dehghani, Minderer, Heigold, Gelly, Uszkoreit, and Houlsby]{vit}
Alexey Dosovitskiy, Lucas Beyer, Alexander Kolesnikov, Dirk Weissenborn, Xiaohua Zhai, Thomas Unterthiner, Mostafa Dehghani, Matthias Minderer, Georg Heigold, Sylvain Gelly, Jakob Uszkoreit, and Neil Houlsby.
\newblock An image is worth 16x16 words: Transformers for image recognition at scale.
\newblock In \emph{International Conference on Learning Representations}, 2021.

\bibitem[Esser et~al.(2021)Esser, Rombach, and Ommer]{Esser2021TamingTransformers}
Patrick Esser, Robin Rombach, and Bj{\"o}rn Ommer.
\newblock Taming transformers for high-resolution image synthesis.
\newblock In \emph{CVPR}, pages 12873--12883, 2021.

\bibitem[Gemini~Team(2025)]{comanici2025gemini25pushingfrontier}
Google Gemini~Team.
\newblock Gemini 2.5: Pushing the frontier with advanced reasoning, multimodality, long context, and next generation agentic capabilities, 2025.

\bibitem[H{\"a}ne et~al.(2020)H{\"a}ne, Tulsiani, and Malik]{Hane2020HSP}
Christian H{\"a}ne, Shubham Tulsiani, and Jitendra Malik.
\newblock Hierarchical surface prediction for 3d object reconstruction.
\newblock \emph{IEEE Transactions on Pattern Analysis and Machine Intelligence}, 42\penalty0 (6):\penalty0 1348--1361, 2020.

\bibitem[Heusel et~al.(2017)Heusel, Ramsauer, Unterthiner, Nessler, and Hochreiter]{fid}
Martin Heusel, Hubert Ramsauer, Thomas Unterthiner, Bernhard Nessler, and Sepp Hochreiter.
\newblock Gans trained by a two time-scale update rule converge to a local nash equilibrium.
\newblock \emph{Advances in neural information processing systems}, 30, 2017.

\bibitem[Ho et~al.(2020)Ho, Jain, and Abbeel]{ddpm}
Jonathan Ho, Ajay Jain, and Pieter Abbeel.
\newblock Denoising diffusion probabilistic models.
\newblock \emph{Advances in neural information processing systems}, 33:\penalty0 6840--6851, 2020.

\bibitem[Hoppe(1996)]{Hoppe1996ProgressiveMeshes}
Hugues Hoppe.
\newblock Progressive meshes.
\newblock In \emph{Proceedings of the 23rd Annual Conference on Computer Graphics and Interactive Techniques (SIGGRAPH '96)}, pages 99--108, New York, NY, USA, 1996. ACM.

\bibitem[Ilharco et~al.(2021)Ilharco, Wortsman, Wightman, Gordon, Carlini, Taori, Dave, Shankar, Namkoong, Miller, Hajishirzi, Farhadi, and Schmidt]{openclip}
Gabriel Ilharco, Mitchell Wortsman, Ross Wightman, Cade Gordon, Nicholas Carlini, Rohan Taori, Achal Dave, Vaishaal Shankar, Hongseok Namkoong, John Miller, Hannaneh Hajishirzi, Ali Farhadi, and Ludwig Schmidt.
\newblock Openclip, 2021.
\newblock If you use this software, please cite it as below.

\bibitem[Jose et~al.(2024)Jose, Moutakanni, Kang, Baldassarre, Darcet, Xu, Li, Szafraniec, Ramamonjisoa, Oquab, Siméoni, Vo, Labatut, and Bojanowski]{dinov2}
Cijo Jose, Théo Moutakanni, Dahyun Kang, Federico Baldassarre, Timothée Darcet, Hu Xu, Daniel Li, Marc Szafraniec, Michaël Ramamonjisoa, Maxime Oquab, Oriane Siméoni, Huy~V. Vo, Patrick Labatut, and Piotr Bojanowski.
\newblock Dinov2 meets text: A unified framework for image- and pixel-level vision-language alignment, 2024.

\bibitem[Kusupati et~al.(2022)Kusupati, Bhatt, Rege, Wallingford, Sinha, Ramanujan, Howard-Snyder, Chen, Kakade, Jain, et~al.]{kusupati2022matryoshka}
Aditya Kusupati, Gantavya Bhatt, Aniket Rege, Matthew Wallingford, Aditya Sinha, Vivek Ramanujan, William Howard-Snyder, Kaifeng Chen, Sham Kakade, Prateek Jain, et~al.
\newblock Matryoshka representation learning.
\newblock \emph{Advances in Neural Information Processing Systems}, 35:\penalty0 30233--30249, 2022.

\bibitem[Labs(2024)]{flux2024}
Black~Forest Labs.
\newblock Flux.
\newblock \url{https://github.com/black-forest-labs/flux}, 2024.

\bibitem[Li et~al.(2024)Li, Tian, Li, Deng, and He]{ar_diff_loss}
Tianhong Li, Yonglong Tian, He Li, Mingyang Deng, and Kaiming He.
\newblock Autoregressive image generation without vector quantization.
\newblock In \emph{Advances in Neural Information Processing Systems}, pages 56424--56445. Curran Associates, Inc., 2024.

\bibitem[Li et~al.(2025)Li, Zhang, Sun, Qi, Li, Cheng, Cai, Wu, Liu, Wang, et~al.]{li2025step1x}
Weiyu Li, Xuanyang Zhang, Zheng Sun, Di Qi, Hao Li, Wei Cheng, Weiwei Cai, Shihao Wu, Jiarui Liu, Zihao Wang, et~al.
\newblock Step1x-3d: Towards high-fidelity and controllable generation of textured 3d assets.
\newblock \emph{arXiv preprint arXiv:2505.07747}, 2025.

\bibitem[Nash et~al.(2020)Nash, Ganin, Eslami, and Battaglia]{polygen}
Charlie Nash, Yaroslav Ganin, S.~M.~Ali Eslami, and Peter~W. Battaglia.
\newblock Polygen: an autoregressive generative model of 3d meshes.
\newblock In \emph{Proceedings of the 37th International Conference on Machine Learning}. JMLR.org, 2020.

\bibitem[Oord et~al.(2018)Oord, Li, and Vinyals]{infonce}
Aaron van~den Oord, Yazhe Li, and Oriol Vinyals.
\newblock Representation learning with contrastive predictive coding.
\newblock \emph{arXiv preprint arXiv:1807.03748}, 2018.

\bibitem[Park et~al.(2019)Park, Kim, Lu, and Cho]{park2019relational}
Wonpyo Park, Dongju Kim, Yan Lu, and Minsu Cho.
\newblock Relational knowledge distillation.
\newblock In \emph{Proceedings of the IEEE/CVF conference on computer vision and pattern recognition}, pages 3967--3976, 2019.

\bibitem[Peebles and Xie(2023)]{scalableDiT}
William Peebles and Saining Xie.
\newblock Scalable diffusion models with transformers.
\newblock In \emph{Proceedings of the IEEE/CVF international conference on computer vision}, pages 4195--4205, 2023.

\bibitem[Radford et~al.(2021)Radford, Kim, Hallacy, Ramesh, Goh, Agarwal, Sastry, Askell, Mishkin, Clark, Krueger, and Sutskever]{Radford2021LearningTV}
Alec Radford, Jong~Wook Kim, Chris Hallacy, A. Ramesh, Gabriel Goh, Sandhini Agarwal, Girish Sastry, Amanda Askell, Pamela Mishkin, Jack Clark, Gretchen Krueger, and Ilya Sutskever.
\newblock Learning transferable visual models from natural language supervision.
\newblock In \emph{ICML}, 2021.

\bibitem[Rippel et~al.(2014)Rippel, Gelbart, and Adams]{nested}
Oren Rippel, Michael Gelbart, and Ryan Adams.
\newblock Learning ordered representations with nested dropout.
\newblock In \emph{International Conference on Machine Learning}, pages 1746--1754. PMLR, 2014.

\bibitem[Samet(1984)]{Samet1984QuadtreeSurvey}
Hanan Samet.
\newblock The quadtree and related hierarchical data structures.
\newblock \emph{ACM Computing Surveys}, 16\penalty0 (2):\penalty0 187--260, 1984.

\bibitem[Siddiqui et~al.(2024)Siddiqui, Alliegro, Artemov, Tommasi, Sirigatti, Rosov, Dai, and Nie{\ss}ner]{meshgpt}
Yawar Siddiqui, Antonio Alliegro, Alexey Artemov, Tatiana Tommasi, Daniele Sirigatti, Vladislav Rosov, Angela Dai, and Matthias Nie{\ss}ner.
\newblock Meshgpt: Generating triangle meshes with decoder-only transformers.
\newblock In \emph{Proceedings of the IEEE/CVF conference on computer vision and pattern recognition}, pages 19615--19625, 2024.

\bibitem[Su et~al.(2024)Su, Ahmed, Lu, Pan, Bo, and Liu]{su2024roformer}
Jianlin Su, Murtadha Ahmed, Yu Lu, Shengfeng Pan, Wen Bo, and Yunfeng Liu.
\newblock Roformer: Enhanced transformer with rotary position embedding.
\newblock \emph{Neurocomputing}, 568:\penalty0 127063, 2024.

\bibitem[Sun et~al.(2024)Sun, Jiang, Chen, Zhang, Peng, Luo, and Yuan]{llamagen}
Peize Sun, Yi Jiang, Shoufa Chen, Shilong Zhang, Bingyue Peng, Ping Luo, and Zehuan Yuan.
\newblock Autoregressive model beats diffusion: Llama for scalable image generation.
\newblock \emph{arXiv preprint arXiv:2406.06525}, 2024.

\bibitem[Vallet and L{\'e}vy(2008)]{Vallet2008ManifoldHarmonics}
Bruno Vallet and Bruno L{\'e}vy.
\newblock Spectral geometry processing with manifold harmonics.
\newblock \emph{Computer Graphics Forum}, 27\penalty0 (2):\penalty0 251--260, 2008.

\bibitem[Wang et~al.(2017)Wang, Liu, Guo, Sun, and Tong]{Wang2017_OCNN}
Peng{-}Shuai Wang, Yang Liu, Yu{-}Xiao Guo, Chun{-}Yu Sun, and Xin Tong.
\newblock O{-}cnn: Octree-based convolutional neural networks for 3d shape analysis.
\newblock \emph{ACM Transactions on Graphics}, 36\penalty0 (4):\penalty0 72:1--72:11, 2017.

\bibitem[Wang et~al.(2024)Wang, Lorraine, Wang, Su, Zhu, Fidler, and Zeng]{wang2024llamameshunifying3dmesh}
Zhengyi Wang, Jonathan Lorraine, Yikai Wang, Hang Su, Jun Zhu, Sanja Fidler, and Xiaohui Zeng.
\newblock Llama-mesh: Unifying 3d mesh generation with language models, 2024.

\bibitem[Wei et~al.(2025)Wei, Wang, Zhou, Chen, and Wang]{octgpt}
Si-Tong Wei, Rui-Huan Wang, Chuan-Zhi Zhou, Baoquan Chen, and Peng-Shuai Wang.
\newblock Octgpt: Octree-based multiscale autoregressive models for 3d shape generation.
\newblock In \emph{Proceedings of the Special Interest Group on Computer Graphics and Interactive Techniques Conference Conference Papers}, pages 1--11, 2025.

\bibitem[Wen et~al.(2025)Wen, Zhao, Elezi, Deng, and Qi]{semanticist}
Xin Wen, Bingchen Zhao, Ismail Elezi, Jiankang Deng, and Xiaojuan Qi.
\newblock " principal components" enable a new language of images.
\newblock \emph{ICCV}, 2025.

\bibitem[Wu et~al.(2024)Wu, Lin, Zhang, Zeng, Xu, Torr, Cao, and Yao]{direct3d}
Shuang Wu, Youtian Lin, Feihu Zhang, Yifei Zeng, Jingxi Xu, Philip Torr, Xun Cao, and Yao Yao.
\newblock Direct3d: Scalable image-to-3d generation via 3d latent diffusion transformer.
\newblock \emph{Advances in Neural Information Processing Systems}, 37:\penalty0 121859--121881, 2024.

\bibitem[Xiang et~al.(2025)Xiang, Lv, Xu, Deng, Wang, Zhang, Chen, Tong, and Yang]{trellis}
Jianfeng Xiang, Zelong Lv, Sicheng Xu, Yu Deng, Ruicheng Wang, Bowen Zhang, Dong Chen, Xin Tong, and Jiaolong Yang.
\newblock Structured 3d latents for scalable and versatile 3d generation.
\newblock In \emph{Proceedings of the Computer Vision and Pattern Recognition Conference}, pages 21469--21480, 2025.

\bibitem[Ye et~al.(2025)Ye, Wang, Zhao, Xie, and Zhu]{ye2025shapellm}
Junliang Ye, Zhengyi Wang, Ruowen Zhao, Shenghao Xie, and Jun Zhu.
\newblock Shapellm-omni: A native multimodal llm for 3d generation and understanding.
\newblock \emph{arXiv preprint arXiv:2506.01853}, 2025.

\bibitem[Yu et~al.(2023)Yu, Lezama, Gundavarapu, Versari, Sohn, Minnen, Cheng, Birodkar, Gupta, Gu, Hauptmann, Gong, Yang, Essa, Ross, and Jiang]{Yu2023MAGVITv2}
Lijun Yu, José Lezama, Nitesh~B. Gundavarapu, Luca Versari, Kihyuk Sohn, David Minnen, Yong Cheng, Vighnesh Birodkar, Agrim Gupta, Xiuye Gu, Alexander~G. Hauptmann, Boqing Gong, Ming-Hsuan Yang, Irfan Essa, David~A. Ross, and Lu Jiang.
\newblock Language model beats diffusion -- tokenizer is key to visual generation.
\newblock \emph{arXiv preprint arXiv:2310.05737}, 2023.

\bibitem[Yu et~al.(2024)Yu, Weber, Deng, Shen, Cremers, and Chen]{tiktok}
Qihang Yu, Mark Weber, Xueqing Deng, Xiaohui Shen, Daniel Cremers, and Liang-Chieh Chen.
\newblock An image is worth 32 tokens for reconstruction and generation.
\newblock \emph{Advances in Neural Information Processing Systems}, 37:\penalty0 128940--128966, 2024.

\bibitem[Yu et~al.(2025)Yu, Kwak, Jang, Jeong, Huang, Shin, and Xie]{repa}
Sihyun Yu, Sangkyung Kwak, Huiwon Jang, Jongheon Jeong, Jonathan Huang, Jinwoo Shin, and Saining Xie.
\newblock Representation alignment for generation: Training diffusion transformers is easier than you think.
\newblock In \emph{The Thirteenth International Conference on Learning Representations}, 2025.

\bibitem[Zhang et~al.(2023)Zhang, Tang, Nie\ss{}ner, and Wonka]{shape2vecset}
Biao Zhang, Jiapeng Tang, Matthias Nie\ss{}ner, and Peter Wonka.
\newblock 3dshape2vecset: A 3d shape representation for neural fields and generative diffusion models.
\newblock \emph{ACM Trans. Graph.}, 42\penalty0 (4), 2023.

\bibitem[Zhang et~al.(2025)Zhang, Siddiqui, Avetisyan, Xie, Engel, and Howard-Jenkins]{vertexregen}
Xiang Zhang, Yawar Siddiqui, Armen Avetisyan, Chris Xie, Jakob Engel, and Henry Howard-Jenkins.
\newblock Vertexregen: Mesh generation with continuous level of detail.
\newblock \emph{ICCV}, 2025.

\end{thebibliography}
 }

\maketitlesupplementary
\section{Additional Qualitative Results}
We provide an extensive gallery of qualitative results in the accompanying \textbf{\emph{supplemental \webpage}}, illustrating the capabilities of our tokenizer and autoregressive (AR) model. These 3D visualizations demonstrate that our method produces high-fidelity reconstructions that visually surpass recent baselines. Note our AR model's flexibility in generating complete and plausible 3D shapes even when conditioned on a few tokens. These qualitative findings are consistent with the strong quantitative performance reported in Table 2 of the main manuscript, further validating the effectiveness of our semantic tokenization strategy.

\section{Shape Retrieval using RIDA}
To quantitatively validate that our Relational Inter-Distance Alignment (RIDA) objective successfully reorganizes the 3D latent space according to semantic salience rather than just geometric proximity, we evaluate our method on a shape retrieval task. Since RIDA is designed to distill the semantic topology of the DINOv2~\cite{dinov2} (\emph{teacher}) space into 3D triplanes, we utilize DINO similarity as the ground truth for defining semantic neighbors.

We compare our RIDA-aligned features against a baseline of (i) raw triplane latents, which primarily capture geometric spatial structure and (ii) a Direct Regression baseline, trained to predict DINO features via explicit supervision. To assess generalization, we conduct this evaluation on two distinct datasets: (i) our In-Distribution set, consisting of held-out samples from our training distribution, and (ii) our Evaluation Set (Out-of-Distribution), which contains shapes generated by the unseen Step1X-3D model with a different underlying representation as specified in Section 4.1 in the manuscript. For the Direct Regression baseline, we employ the same transformer backbone as RIDA but replace the relational contrastive and distillation objectives with a direct spatial regression loss (MSE). Empirically, we observe that this direct mapping is ineffective; the network suffers from optimization stagnation, exhibiting early loss plateauing on the validation set and failing to capture discriminative semantic features.

We report Recall@K to measure the proportion of ground-truth semantic neighbors successfully retrieved, and Mean Average Precision (mAP@K) to evaluate the quality of their ranking within the top results. Additionally, we compute the Jaccard Index to quantify the set intersection-over-union (IoU) between the retrieved candidates and the ground truth.
 
\begin{table}[t!]
\centering
\caption{\textbf{Shape Retrieval Evaluation.} We evaluate RIDA against raw triplane latents and a direct regression baseline that is trained to predict DINOv2 features. Ground truth neighbors are defined by DINOv2 similarity. RIDA (ours) outperforms the geometric baseline on both the in-distribution validation set and the out-of-distribution evaluation set (generated using Step1X-3D ), confirming that RIDA effectively captures semantics.}
\label{tab:retrieval}
\resizebox{\columnwidth}{!}{%
\begin{tabular}{l|ccc|ccc}
\toprule
 & \multicolumn{3}{c|}{\textbf{Out-of-Distribution Set}} & \multicolumn{3}{c}{\textbf{In-Distribution Set}} \\
\textbf{Metric} & \textbf{Triplane} & \textbf{Regression} & \textbf{RIDA (ours)} & \textbf{Triplane} & \textbf{Regression} & \textbf{RIDA (ours)} \\
\midrule
\multicolumn{5}{l}{\textit{Top-3 Retrieval (K=3)}} \\
Recall@3 & 20.20\% & 20.63\% & \textbf{32.03\%} & 19.07\% & 27.00\% & \textbf{48.50\%} \\
mAP@3 & 17.47\% & 17.28\% & \textbf{28.28\%} & 16.42\% & 22.90\% & \textbf{44.28\%} \\
Jaccard & 13.73\% & 13.91\% & \textbf{23.25\%} & 13.60\% & 19.19\% & \textbf{38.36\%} \\
\midrule
\multicolumn{5}{l}{\textit{Top-5 Retrieval (K=5)}} \\
Recall@5 & 19.54\% & 20.70\% & \textbf{30.30\%} & 18.48\% & 26.80\% & \textbf{47.90\%} \\
mAP@5 & 15.12\% & 15.29\% & \textbf{24.71\%} & 14.42\% & 21.16\% & \textbf{41.85\%} \\
Jaccard & 12.42\% & 13.22\% & \textbf{20.49\%} & 12.01\% & 17.80\% & \textbf{35.77\%} \\
\bottomrule
\end{tabular}%
}
\end{table}

\begin{table*}[t]
    \centering %
    \caption{\textbf{Ablation tokenizer reconstruction.} We compare LoST trained without the proposed RIDA semantic alignment across multiple decoding levels. RIDA consistently improves semantic reconstruction quality, as measured by DINO and DINOv2 similarity, with the largest gains appearing in the low-token regime where semantic guidance is most critical. This confirms that RIDA helps short token prefixes encode more semantically meaningful structure. The best score at each decoding level is highlighted in \textbf{bold}.}
    \label{tab:ablation}
    
    \resizebox{0.7\textwidth}{!}{
        \begin{tabular}{r | rrrrr | rrrrr}
            \toprule
            
            {} 
            & \multicolumn{5}{c}{w/o RIDA} 
            & \multicolumn{5}{c}{\textbf{w/ RIDA (ours)}} \\
            
            \cmidrule(lr){2-6} \cmidrule(lr){7-11} 
            
            Num Tokens $\rightarrow$ & \textbf{1} & \textbf{4} & \textbf{16} & \textbf{64} & \textbf{512}
            & \textbf{1} & \textbf{4} & \textbf{16} & \textbf{64} & \textbf{512} \\
            
            \midrule

            DINO$\uparrow$
            & 0.720 & 0.758 & \textbf{0.821} & 0.876 & 0.904
            & \textbf{0.731} & \textbf{0.765} & 0.814 & \textbf{0.880} & \textbf{0.921} \\

            DINOv2$\uparrow$
            & 0.528 & 0.590 & 0.693 & 0.763 & 0.867
            & \textbf{0.556} & \textbf{0.612} &\textbf{ 0.694} & \textbf{0.805} & \textbf{0.875} \\
            
            \bottomrule
        \end{tabular}
    } %

\end{table*}

As shown in Table~\ref{tab:retrieval}, RIDA demonstrates superior semantic alignment compared to the geometric baseline. On the challenging OOD evaluation set , our method significantly improves Mean Average Precision (mAP@3) from 17.47\% to 28.28\%, proving that RIDA captures abstract semantic identity that is robust to low-level geometric variations. This performance gap is further amplified on the in-distribution set (benefiting from the same VAE latent encoding), where RIDA achieves a 44.28\% mAP compared to the baseline's 16.42\% (triplane) and 22.90\% (feature regression). These results confirm that while raw triplanes are limited to geometric matching, RIDA effectively aligns 3D shapes with the rich semantic hierarchy of DINO. The results on the in-distribution set are critical for training \name.  RIDA also significantly surpasses our direct regression baseline that is trained to predict DINOv2 features.

\section{Ablation on RIDA}
Following recent advances in image tokenization~\cite{flextok, semanticist}, which leverage alignment losses like REPA~\cite{repa} to structure latent spaces, we integrate RIDA to explicitly align our 3D triplane representations with semantic priors. While the diffusion decoder possesses an inherent capacity to move towards plausible geometry via its generative prior—especially at lower guidance scales—we find that RIDA significantly augments this capability. By enforcing a structured relationship between the 3D latent space and the teacher's semantic embedding, RIDA serves as a potent regularizer that enhances the semantic consistency of the decoded shapes. Furthermore, the semantic supervision acts as a stabilizing factor, counteracting the inherent training volatility introduced by the nested dropout mechanism.

This benefit is most pronounced in low-bitrate regimes. As demonstrated in \Cref{tab:ablation} particularly DINOv2 similarity scores and FID, when the token budget is constrained, the model cannot rely solely on dense geometric encoding. In these settings, RIDA effectively bridges the gap between high-level semantic intent and geometric reconstruction, yielding substantial quantitative gains and ensuring that even short token prefixes decode into semantically recognizable structures. The Chamfer Distance remains similar in both settings, which suggests that utilizing RIDA does not negatively impact training for geometry but enhances semantic alignment. We further note that extended training of the diffusion decoder eventually leads to convergence without RIDA, our method accelerates this process ($\sim$40\% faster). These findings are consistent with Flextok's ablation study~\cite{flextok} when using REPA.

Note our ablation of a direct regression baseline that is trained to predict DINO features directly in \Cref{tab:retrieval}; this approach fails to accurately learn semantics. We present qualitative results on the ablation in the \textbf{supplemental \webpage}.

\section{Details about RIDA}

\paragraph{Inter-Instance Rank Distillation.}
The contrastive loss enforces separation based on hard thresholds between positive and negative samples, but discards the rich, continuous relational structure within the teacher's space. 
This continuous structure is essential, but it is non-trivial to transfer to the student space. 
To this end, we are inspired by Relational Knowledge Distillation (RKD)~\cite{park2019relational}, which transfers pairwise Euclidean distances. In our setting, we use cosine similarities 
$\mathbf{c}^s_i \coloneqq [c_{ij}]_{z_j \in \mathcal{B}\ \text{with}\ i\neq j}$
and the corresponding similarities in the teacher space $\mathbf{c}^t_i$.
However, in our cross-modal setting (3D-to-2D), absolute similarities are not directly comparable; a naive loss on raw cosine similarities ($\|\mathbf{c}^s_i - \mathbf{c}^t_i\|_2^2$), fails to converge, as the student and teacher's per-anchor similarity distributions (i.e., their means $\mu(\cdot)$ and scales $\sigma(\cdot)$) are fundamentally misaligned.
We therefore introduce a rank distillation loss, which is designed to be \emph{invariant} to these modality-specific affine transformations. Instead of matching individual pairs, we match the \emph{entire} per-anchor similarity vector. We achieve invariance by standardizing (z-scoring) each anchor's similarity row independently to remove its specific mean and scale, thus isolating the pure relational pattern:

\begin{equation}
\label{eq:z_score}
\widetilde{\mathbf{c}}^s_i = \frac{\mathbf{c}^s_i - \mu(\mathbf{c}^s_i)}{\sigma(\mathbf{c}^s_i)},
\qquad
\widetilde{\mathbf{c}}^t_i = \frac{\mathbf{c}^t_i - \mu(\mathbf{c}^t_i)}{\sigma(\mathbf{c}^t_i)}.
\end{equation}
The loss is the Mean Squared Error between these z-scored, distribution-invariant vectors:
\begin{equation}
\label{eq:rid}
\mathcal{L}_{\mathrm{rank}}
:= \mathbb{E}_{\mathbf{z}_i \in \mathcal{B}} \left[ \left\| \widetilde{\mathbf{c}}^s_i - \widetilde{\mathbf{c}}^t_i \right\|^2_2 \right].
\end{equation}
This objective is mathematically proportional to maximizing the Pearson correlation coefficient for each row, as $\|\widetilde{a}-\widetilde{b}\|_2^2 \propto (1-\mathrm{corr}(a,b))$ for z-scored vectors $\widetilde{a}$ and $\widetilde{b}$. By factoring out the per-anchor mean and standard deviation, $\mathcal{L}_{\mathrm{rank}}$ purely optimizes for the \emph{relative neighborhood ranking}, which is the core semantic relation we distill.

\paragraph{Spatial Structure Distillation.}
To ensure the student's spatial tokens \(\mathbf{S}^s_i\) capture the same part-level relationships as the teacher's \(\mathbf{S}^t_i\), we distill the intra-instance token affinities. Instead of a direct L2 match, we match the \emph{distribution} of similarities. We compute self-affinity matrices $\mathbf{A}^s_i, \mathbf{A}^t_i \in \mathbb{R}^{K \times K}$ within each instance $i$ by:
\begin{equation}
\label{eq:affinity}
\mathbf{A}_i[k, \ell] = \langle \mathbf{S}_{i,k}, \mathbf{S}_{i,\ell} \rangle. 
\end{equation}
 We then apply a row-wise softmax to create affinity distributions, $\mathbf{a}_{i,k} = \text{Softmax}(\mathbf{A}_i[k, \cdot])$. The spatial loss minimizes the KL divergence between the teacher and student distributions for each token:
\begin{equation}
\label{eq:spatial_kl}
\mathcal{L}_{\mathrm{spatial}} := \mathbb{E}_{i, k} \left[ D_{\mathrm{KL}}\big( \mathbf{a}^t_{i,k} \,\|\, \mathbf{a}^s_{i,k} \big) \right].
\end{equation}
This forces the student's tokens to learn the same relative 
affinity
patterns as the teacher's, preserving local geometric structure.

The final semantic pretraining objective for our student encoder \(f_\theta\) is a weighted sum of these components:
\begin{equation}
\label{eq:total_sem_loss}
\mathcal{L}_{\mathrm{RIDA}}
:= \lambda_{\mathrm{g}}\mathcal{L}_{\mathrm{global}}
+ \lambda_{\mathrm{r}}\mathcal{L}_{\mathrm{rank}}
+ \lambda_{\mathrm{s}}\mathcal{L}_{\mathrm{spatial}}.
\end{equation}
We use $\lambda_{\mathrm{global}}=1.0$, $\lambda_{\mathrm{rank}}=1.0$, and $\lambda_{\mathrm{spatial}}=0.5$ in our experiments. The resulting network \(f_\theta\) provides a semantically-structured 3D latent space, which we can now leverage as a powerful loss function for our generative task.

\section{Extending \name to other 3D Representations}
\begin{figure}[h]
   \centering
  \includegraphics[width=0.8\columnwidth]{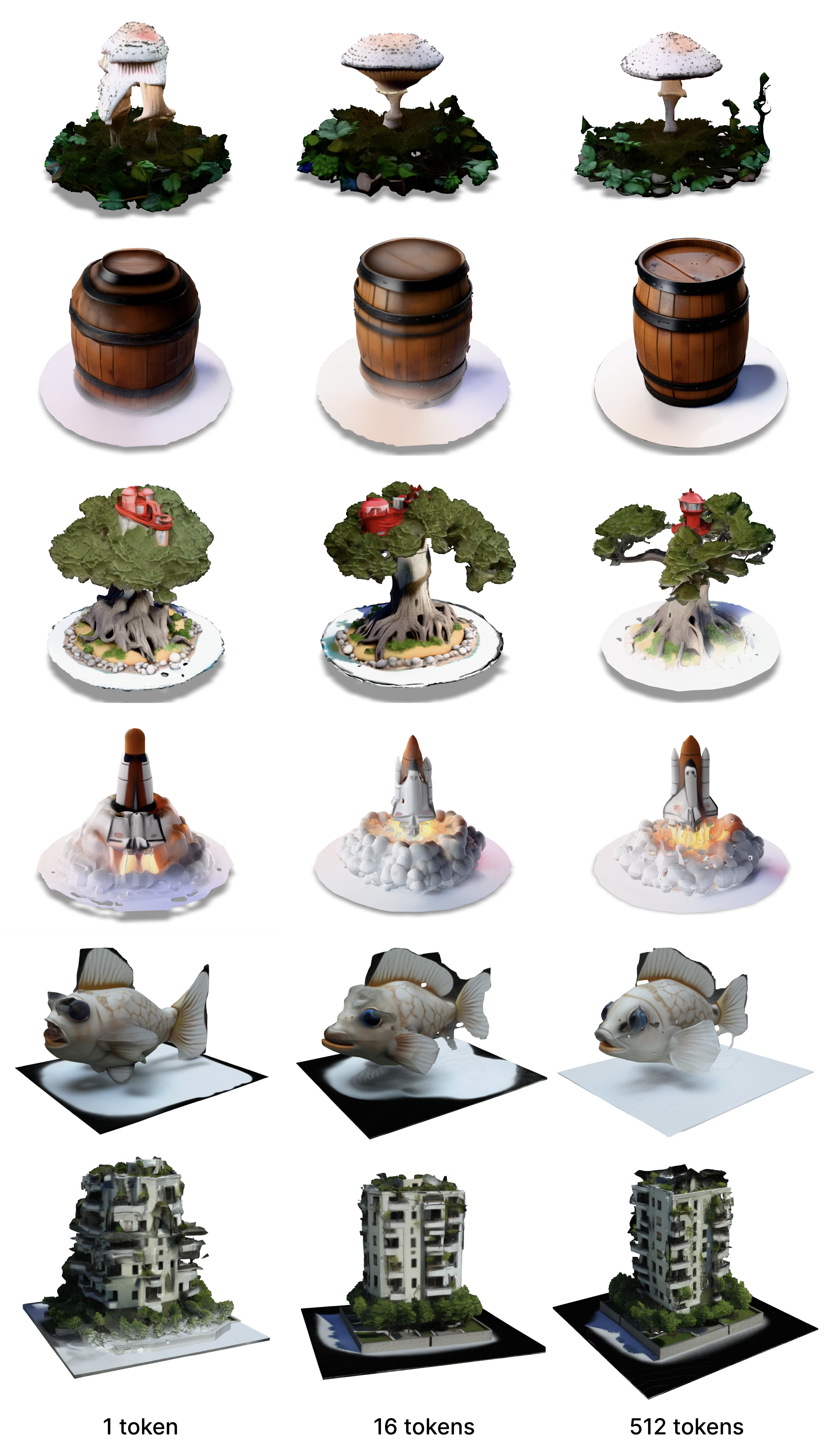}
   \caption{\textbf{\name in the TRELLIS 3D VAE latent space}. LoST applied to TRELLIS Stage-1 latents demonstrate that the proposed tokenization generalizes beyond the Direct3D triplane representation. These results highlight the flexibility of LoST as a representation-agnostic tokenizer for variable-length 3D generation. }
   \label{fig:trellis}
\end{figure}

Our \name framework and RIDA objective are designed to be representation-agnostic. Here, we showcase \name in the TRELLIS~\cite{trellis} latent space. We operate on TRELLIS Stage-1 latents by reshaping the $16^3$ voxel grid (feature dimension 8) into a $64^2$ 2D grid, which preserves our original architecture with 16-dimensional register tokens similar to our adaptation of Direct3D's triplanes. Our method produces variable-length representations that are decoded via TRELLIS Stage-2 to recover high-frequency details, similar to ShapeLLM-Omni~\cite{ye2025shapellm}, while supporting a flexible token budget. We present qualitative results and quantitative results for tokenization on the Objaverse dataset~\cite{deitke2023objaverse} in \Cref{fig:trellis} and \Cref{tab:objaversetokenizer} respectively. TRELLIS additionally models texture compared to Direct3D, which focuses on geometry alone. These results validate the generalization and flexibility of our method.

\newcommand{\best}[1]{\textbf{#1}}
\newcommand{\second}[1]{\underline{#1}}
\begin{table}[h]
\caption{Tokenization Evaluation on Objaverse. We evaluated 128 high-quality watertight Objaverse assets (filtered via Step1X-3D). These results are consistent with our results on our evaluation set in ~\Cref{tab:tokenizer}. We note that all evaluation is computed on untextured renderings, which focuses on geometry alone (best results are in bold, second best are underlined).}
    \centering\resizebox{\columnwidth}{!}{
        \begin{tabular}{rccc}
            \toprule
             & Num Tokens & CD ($\times 10^{-2}$)$\downarrow$ & DINO$\uparrow$ \\
            \midrule
            \multirow{3}{*}{OctGPT}
            & $\sim$219    & 17.925 & 0.529 \\
            & $\sim$15,031  & \second{1.210} & 0.611 \\
            & $\sim$239,004 & \best{0.123} & 0.729 \\
            \midrule
            \multirow{3}{*}{VertexRegen} 
            & $\sim$3,521 & 3.612 & 0.545 \\
            & $\sim$3,701 & 1.593 & 0.595 \\
            & $\sim$8,321 & 0.625 & 0.753 \\
            \midrule
            \multirow{3}{*}{\textbf{LoST (Direct3D)}}
            & 1   & \best{2.460} & \best{0.690} \\
            & 16  & \best{0.963} & \best{0.779} \\
            & 512 & 0.385 & \best{0.874} \\
            \midrule
            \multirow{3}{*}{\textbf{LoST (Trellis)}} 
            & 1   & \second{3.242} & \second{0.631}\\
            & 16  & 1.351 & \second{0.702}\\
            & 512 & \second{0.345} & \second{0.801} \\
            \bottomrule
        \end{tabular}
    }
    \label{tab:objaversetokenizer}
\end{table}

\begin{table}[h]
\vspace{-0.5em}
\small
\centering
\setlength{\tabcolsep}{4.5pt}
\caption{We compare total token dimension cost against other autoregressive methods. We note ShapeLLM-Omni uses 32-dimensional tokens for representation but this increases to 3584 due to LLM usage.}
\resizebox{\columnwidth}{!}{
\begin{tabular}{r|ccc}
& Num Tokens & Token Dimension & Total \\
\hline
OctGPT~\cite{octgpt} &       $\sim$ 50,000         &      1 & 50,000          \\
Llama-Mesh~\cite{wang2024llamameshunifying3dmesh}  &             $\sim$ 3758    &       4096   &   $\sim$ 15,392,768    \\
ShapeLLM-Omni~\cite{ye2025shapellm}  &             1024     &       3584    &  3,670,016    \\
MeshGPT~\cite{meshgpt} &  1200 - 4800 & 192 & 230,400 - 921,600 \\
LoST GPT (ours) &             \textbf{128}     &       \textbf{32}   &    \textbf{4,096}    \\
\bottomrule
\end{tabular}}
\label{table:dimensions}
\end{table}

\section{Further Implementation Details}
\paragraph{Evaluation.} 
We render four orthogonal views per shape using Blender with detailed shading. We compute all perceptual metrics for each view and report the averaged results.
\paragraph{Tokenizer Training Details.} 
We train the initial 50 epochs without nested dropout to allow the model to prioritize shape reconstruction using its full capacity, while retaining causal masking throughout. We employ mixed precision training with `bf16' and utilize Exponential Moving Average (EMA) for model weight updates to stabilize training. While we did not explore learned positional encodings or RoPE~\cite{su2024roformer}, incorporating these could potentially yield further performance gains. We use an effective batch size of 256 across 8 GPUs for training \name. 
\paragraph{Text Prompt.} We provide the prompt template used in Gemini2.5 Pro~\cite{comanici2025gemini25pushingfrontier} to produce prompts used to generate our dataset in the next page. In each API call to Gemini, we only produce 500 prompts at a time to ensure the highest quality.

\clearpage
\newpage
\begin{promptbox}
You are a highly creative and meticulous prompt generator for a cutting-edge text-to-3D diffusion model. Your primary task is to generate **500 unique text prompts**, each describing a **single, distinct, and highly visual 3D object or structured scene element.**

---

Goal and Expansive Diversity Constraints:

The generated collection of objects must be **hyper-varied** and **maximally diverse**, drawing inspiration from all forms of media, history, and imagination. Ensure the prompts comprehensively cover the following major categories, with rich, descriptive detail:

1.  **Everyday, Tools, and Artifacts:**
    * **Practical:** A perfectly arranged sushi bento box, a complex wind-up clock mechanism, an antique brass telescope.
    * **Relics \& Treasures:** A glowing Atlantean crystal, a ceremonial Mayan mask, a cursed dagger encrusted with jewels.
    * Accessories, jewelry, gadgets, household items, musical instruments, sports equipment, clothes, office supplies, toys, etc, etc. Endless possibilties.
2.  **Characters, Creatures, and Figurines:**
    * **Characters from popular culture:**: Examples: Anime characters, Marvel characters such as spider-man, hulk, etc, characters from games such as Ezio Auditore, Lara Croft, Mario, Kratos, cartoon characters, etc., Indian Jones, Sherlock Holmes, Harry Potter, human characters, standard humans such as man, woman, kid, etc.
    * **Fantasy \& Sci-Fi:** Intricate elves, biomechanical cyborgs, ethereal spirits, Lovecraftian monsters.
    * **Pop Culture \& History:** cinematic creatures in dynamic poses, stylized political figures, classic literary characters, characters from animations, sports,.
    * **Abstract/Stylized:** Chibi characters, low-poly mascots, geometric avatars.
3.  **Architecture, Structures, and Scenics:**
    * **Internal \& External:** A collapsing spiral staircase, a sleek Brutalist building facade, an ornate Victorian greenhouse, a subterranean alien throne room, Taj Mahal, Tokyo Tower, Hanging gardens of babylon, sydney opera house.
    * **Specific Styles:** Hyper-realistic, stylized claymation, cel-shaded, vaporwave aesthetic.
4.  **Vehicles and Machinery:**
    * **Operational \& Conceptual:** Detailed vintage motorcycles, futuristic flying battleships, abandoned industrial robots, specialized scientific equipment (e.g., a particle accelerator component).
    * **Condition:** Rusted, pristine, battle-damaged, overgrown with moss.
5.  **Organic, Flora, and Fauna:**
    * **Animals:** Photorealistic wildlife (e.g., a snow leopard mid-leap), mythical beasts (e.g., a hydra emerging from water), taxidermy displays.
    * **Plants:** Rare succulents, carnivorous plants, an entire ancient, etc.
    * Various fruits and vegetables, flowers, trees, fungi, etc. Don't do bonsai, we already have many bonsai prompts. Rather explore diverse things.
    * Food items and dishes: gourmet dishes, desserts, beverages, noodles, etc.

This is just a guide-- use your creativity to explore and expand upon these categories, do not limit yourself to them. 
** Choose from absolutely random stuff. Keep a mix of realistic everyday objects/things and creative ones. Focus more on realistic/hyperrealistic. Keep very few futuristic items**
---

Formatting and Output Rules:

* Generate **exactly 500** unique, highly visual prompts. **DO NOT REPEAT** any prompt.
* The output must contain **only** the sequentially numbered list of prompts. **DO NOT INCLUDE** any introductory text, conversational fillers, or surrounding markdown/code blocks.
* The numbering must start at **1.** and proceed sequentially. Ensure **each prompt is on a new line**.

**Sample Output (Do NOT repeat these exact prompts):**
1. Porcghe 911 Carrera S, hyperrealistic
2. statute of David
3. a sleek, angular neon sign that reads "VOID"
4. an intricate, highly detailed mechanical dragonfly with copper wings
5. a crumbling statue of a griffin perched on a stone pillar
...
500. superhero in a dynamic pose, highly detailed (you can use various superheroes/popular characters/anime characters/game characters)

\end{promptbox}

\end{document}